\documentclass[lettersize,journal]{IEEEtran}
\usepackage{amsmath,amsfonts}
\usepackage{algorithmic}
\usepackage{algorithm}
\usepackage{array}
\usepackage[caption=false,font=normalsize,labelfont=sf,textfont=sf]{subfig}
\usepackage{textcomp}
\usepackage{stfloats}
\usepackage{url}
\usepackage{verbatim}
\usepackage{graphicx}
\usepackage{cite}
\usepackage{xcolor}
\usepackage{amsmath}
\usepackage{breqn}
\usepackage{multirow}
\usepackage{booktabs}
\usepackage{graphicx}
\usepackage{subcaption}
\usepackage{amsmath}
\usepackage[export]{adjustbox}
\title{A Multitask Deep Learning Model for Classification and Regression of Hyperspectral Images: Application to the large-scale dataset}
\author{Koushikey Chhapariya\textsuperscript{1,2}, Alexandre Benoit\textsuperscript{2}, Krishna Mohan Buddhiraju\textsuperscript{1}, Anil Kumar\textsuperscript{3}
\\{\textsuperscript{1}Centre of Studies in Resources Engineering, Indian Institute of Technology Bombay, Mumbai, India\\\textsuperscript{2}LISTIC, Université Savoie Mont-Blanc, Annecy, FR\\\textsuperscript{3}Indian Institute of Remote Sensing, ISRO, Dehradun, India\thanks{The work of Koushikey Chhapariya was supported by the Indo-French Centre for Promotion of Advanced Research (IFCPAR/ CEFIPRA) under the Raman–Charpak Fellowship program 2022.}}}

\begin{document}

\maketitle

\begin{abstract}

Multitask learning is a widely recognized technique in the field of computer vision and deep learning domain. However, it is still a research question in remote sensing, particularly for hyperspectral imaging. Moreover, most of the research in the remote sensing domain focuses on small and single-task-based annotated datasets, which limits the generalizability and scalability of the developed models to more diverse and complex real-world scenarios. Thus, in this study, we propose a multitask deep learning model designed to perform multiple classification and regression tasks simultaneously on hyperspectral images. We validated our approach on a large hyperspectral dataset called TAIGA, which contains 13 forest variables, including three categorical variables and ten continuous variables with different biophysical parameters. We design a sharing encoder and task-specific decoder network to streamline feature learning while allowing each task-specific decoder to focus on the unique aspects of its respective task.

Additionally, a dense atrous pyramid pooling layer and attention network were integrated to extract multi-scale contextual information and enable selective information processing by prioritizing task-specific features. Further, we computed multitask loss and optimized its parameters for the proposed framework to improve the model performance and efficiency across diverse tasks. A comprehensive qualitative and quantitative analysis of the results shows that the proposed method significantly outperforms other state-of-the-art methods. We trained our model across $10$ seeds/trials to ensure robustness. Our proposed model demonstrates higher mean performance while maintaining lower or equivalent variability. To make the work reproducible, the codes will be available at https://github.com/Koushikey4596/Multitask-Deep-Learning-Model-for-Taiga-datatset.

\end{abstract}


\section{Introduction}
\IEEEPARstart{H}{yperspectral} imaging (HSI) is a remote sensing technique (also known as image spectrometry) consisting of hundreds of contiguous and spectrally narrow wavelength bands. This detailed spectral information helps in distinguishing similar features and has applications in various domains such as agriculture sector~\cite{alahmari2023hybrid, xie2024hyperspectral}, urban target detection~\cite{9941145, gakhar2024extraction}, the mining industry~\cite{zhang2023mapping, dieters2024robot}, environmental and natural resource management~\cite{bioucas2013hyperspectral, alboody2023new}, surveillance~\cite{ uzkent2017aerial, xu2023mapping}, etc. However, the high dimensionality of data increases storage and computational costs. Moreover, the curse of dimensionality~\cite{hughes1968mean} and redundancy in adjacent spectral bands limit the accuracy of HSI classification. Thus, there is a need for automated learning algorithms to reduce the time consumption for manual work and increase overall efficiency. Many machine learning algorithms, such as KNN, SVM, MLP, etc., have performed satisfactorily but require experts to manually tune parameters that fail to handle unknown/unseen and complex scenes. This increases the need for more robust and adaptive learning methods, thus introducing deep learning algorithms with HSI ~\cite{9082155, 10283327}. Convolutional Neural Networks (CNNs) are one such deep learning approach that is widely considered for extracting learned features with its hierarchical properties and layered structure. 

The acquired hyperspectral images can be represented as a data cube with two continuous axes of spatial dimensions and one axis for spectral dimension. Several researchers~\cite{paoletti2019deep, ghamisi2017advanced, 10430700} have demonstrated the importance of using both spectral and spatial information for hyperspectral image analysis to improve image classification accuracy. The 3-dimensional CNN algorithm can extract discriminatory features from the HSI data cube considering its spectral and spatial information. Despite several research works on HSI with CNN, there is still a limitation with the availability of hyperspectral datasets. Moreover, the prediction of continuous variables from HSI using CNN is little known and understudied ~\cite{hamouda2020smart} while the majority of studies in the field have focused on land cover classification, typically approached as a single-task method. Additionally, learning algorithms demand substantial amounts of data before yielding meaningful results. Thus, computational expenses rise when dealing with multiple tasks, each demanding significant computational resources and extensive training~\cite{9955063}.

Multitask learning~\cite{liu2020multitask} aims at solving multiple prediction problems from a given data sample to improve learning efficiency and classification accuracy by sharing information among different tasks. Multitask learning is a well-known method in the computer vision domain, while it is still a research question in the remote sensing domain~\cite{lian2024multitask}. The main motivation for this research is to develop a learning-based algorithm that can enable multitask learning for both regression and classification tasks while maintaining or improving inference speed performances on an independent test set. Despite their distinct objectives, regression and classification tasks in remote sensing share commonalities in supervised learning and are expected to utilize similar learning patterns and techniques. This common ground forms the basis for employing a multitask approach, where a single model is trained to handle both regression and classification tasks simultaneously. While significant research has been conducted on multitask learning models, particularly in the context of image segmentation~\cite{wang2020boundary, cui2023mtscd, liu2024hybrid}, there has been relatively limited exploration of multitask learning within hyperspectral image classification~\cite{9673792}. With this observation, our research aims to address the following research question:
\begin{enumerate}
    \item Is there an existing multitask learning framework capable of simultaneously predicting different classes and computing values for continuous variables specifically designed for large hyperspectral datasets?
    \item How does multitask behave when applied to hyperspectral images, particularly considering the presence of a large number of bands, and what impact does the integration of spectral and spatial information have on its performance?
    \item Are there effective methods available to handle a class imbalance in the data and achieve task balancing to compute the multitask loss and optimize its parameters within a multitask learning framework?
\end{enumerate}

This paper proposes a multitask deep-learning framework for classification and regression for a large hyperspectral dataset to answer the above-mentioned questions. The main focus of our work is to propose a framework that can assign classes to input data and predict continuous outcomes simultaneously. We considered an encoder-decoder-based framework for integrating multiple tasks within a single model and shared representation learning. Further, we append a dilated (atrous) convolutional module to improve the network's contextual reasoning. From a shared latent representation, the multitask decoder is designed, where we first separate the classification and regression tasks into two distinct paths, each being more task-specific. 
Finally, task-specific heads are optimized, relying on a single fully connected layer. The majority of the model’s parameters and depth are in the feature encoding, with very little complexity in each task decoder.  This highlights the advantage of multitask learning, where computational resources can be efficiently shared to learn a more effective shared representation. This work also aims to handle data imbalance, which is commonly observed in large-scale datasets. 
Additionally, task balancing for the multitask learning method and computing balanced loss function for both classification and regression are other challenges addressed by this research, thus producing optimal model convergence and better accuracy compared with learning each task individually. Experimental studies show that the proposed model can achieve excellent classification
results on the given datasets. The main contribution of this paper is summarized as follows.

\begin{enumerate}
    \item We developed a multitask deep learning model capable of simultaneously predicting various classes and computing continuous values for large-scale datasets.
    \item We designed a framework to analyze the behavior of multitask approaches when applied to hyperspectral images characterized by a large number of bands. Additionally, this framework evaluates the impact of integrating both spectral and spatial information on the performance of the multitask model.
    \item We introduce cost-sensitive learning and focal loss methods for handling class imbalance in data and achieving task balancing for computing multitask loss and optimizing its parameters within a multitask learning framework, aiming to enhance model performance and efficiency for different tasks. 
\end{enumerate}

The remaining paper is organized as follows.  Section II provides the background and motivation of the proposed work. Section III explains the dataset used and the proposed model of this research work. Section IV presents the quantitative and qualitative results of the proposed model. Further, Section V concludes the research work with future directions.

\section{Related Work}
In recent decades, many HSI classification and regression frameworks have been proposed for individual tasks due to their broad applications. Traditional methods, such as principal component analysis (PCA) ~\cite{rodarmel2002principal}, linear discriminant analysis (LDA)~\cite{li2011locality}, local linear embedding~\cite{roweis2000nonlinear}, sparse representation~\cite{zhang2016spectral}, manifold learning~\cite{duan2020local}, etc. have been used as feature extraction and dimensionality reduction methods. These methods were proposed to minimize noise and redundancy within the data. However, they require manual parameter tuning by experts and the selection of spectral bands with optimal information content to prevent substantial loss of information. Further, these methods are limited to extracting low-level hand-crafted features, making them inadequate for handling unknown or complex scenes~\cite{li2019deep}. Additionally, challenges such as sensor configurations, atmospheric interference, and spectral variations further complicate HSI data analysis, leading to increased intraclass variability and interclass similarity~\cite{haut2019visual, paoletti2019deep}. Consequently, there's a growing demand for more advanced techniques in HSI classification. Recently, learning-based algorithms, particularly deep learning models, have gained significant attention due to their adaptive nature and significant classification performance. 

Deep Learning models offer the capability to extract high-level abstract features through end-to-end hierarchical frameworks, effectively integrating spectral and spatial information from the input data. Among many deep learning algorithms such as Extended Morphological Profiles (EMPs)~\cite{9883107}, Stacked Autoencoder (SAE)~\cite{zabalza2016novel}, Deep Belief Networks (DBNs)~\cite{chen2020hyperspectral}, Recurrent Neural Networks (RNNs)~\cite{hang2019cascaded}, and Generative Adversarial Networks (GANs) ~\cite{wang2020adaptive}, etc., CNNs has shown efficacy in achieving satisfactory performance for HSI classification. The evolution of CNNs has progressed from one-dimensional (1-D) CNNs ~\cite{hu2015deep}, which extract spectral information, to two-dimensional (2-D) CNNs~\cite{mou2019learning}, incorporating spatial aspects, and eventually to three-dimensional (3-D) CNNs ~\cite{roy2019hybridsn, 10401954}, which simultaneously consider both spectral and spatial information. Although CNNs are generally used for pixel-wise image classification tasks to label different data classes at the image or pixel level, they may not be as effective in predicting continuous variables. In such scenarios, regression-based methods often outperform them~\cite{adhikari2023comparison}.

Many researchers have introduced regression-based methods that aim to estimate the quantitative properties of the scene captured by the hyperspectral data. Some common regression techniques applied to hyperspectral data include Partial Least Squares Regression (PLSR)~\cite{konda2008partial}, Support Vector Regression (SVR)~\cite{axelsson2013hyperspectral}, Gaussian Process Regression (GPR)~\cite{verrelst2016spectral}, and Neural Network Regression (NNR)~\cite{xu2021estimation}. These methods are used for tasks such as estimating physical parameters (e.g., vegetation biomass and soil moisture content), predicting environmental variables (e.g., temperature and humidity), and hyperspectral image reconstruction. Khodadadzadeh et al.~\cite{6822503} introduced a multinomial logistic regression method for hyperspectral classification. The accuracy was further improved by using Kronecker factorization-based multinomial logistic regression algorithm proposed by Wang et al.~\cite{9515683}. However, all these different classification and regression methods have been performed as a single-task method. Using a deep learning approach usually yields more parameters, training data, as well as memory and computational costs to generate significant results. Thus, the computational expenses will increase manyfold if we have to deal with multiple tasks one at a time, each demanding significant computational resources and extensive training. 

Caruana, R.~\cite{caruana1997multitask} laid the foundation for applying the multitask learning approach, where he demonstrated that sharing knowledge across related tasks can improve model performance. Multitask learning was initially introduced in the computer vision domain to mitigate computational challenges by concurrently handling multiple tasks through shared representation techniques, thereby enhancing efficiency and reducing resource demands~\cite{ruder2017overview, vandenhende2020revisiting}. In deep learning, multitask networks are commonly categorized into hard or soft parameter sharing of hidden layers. In the soft-sharing approach, task-specific parameters are assigned to each task, aiding cross-task interaction for feature sharing ~\cite{liu2019end}. Conversely, the hard-sharing method segregates the network into shared and task-specific parameters~\cite{kendall2018multi}. Recently, many methods have been introduced to explore multitask learning in remote sensing applications. Wang et al. ~\cite{wang2020boundary} proposed a novel boundary-aware multitask learning where boundary detection is employed as an auxiliary task to regularize the other two tasks of semantic segmentation and height estimation.
Additionally, Guo et al. ~\cite{9171479} proposed a scene-driven multitask parallel attention convolutional network (MTPA-Net) to fully utilize building extraction from remote sensing imagery. Improving this method, Hong et al. ~\cite{hong2023multi} introduced a multi-task learning network for building extraction and change detection from remote sensing images. This method uses a Swin transformer as a shared backbone network and multiple heads to predict building labels and changes. Liu et al. ~\cite{10409251} proposed a hybrid multitask learning framework (HyMuT) by sharing representations across multiple tasks, which is based on the similarity between the data and the target classification task. Most of these methods aim to improve semantic segmentation performance by combining tasks such as edge detection, building height estimation, change detection, boundary detection, etc. Different from previous remote sensing multitask approaches, our proposed model achieves both classification and regression tasks within a unified framework and further utilizes an automatic task balancing method to compute the multitask loss and to improve their overall performance while reducing the computational costs involved by manual or grid search of the best task weights.




\section{Methodology}
\label{meth}

\begin{figure*}[!h]
  \includegraphics[width=\textwidth]{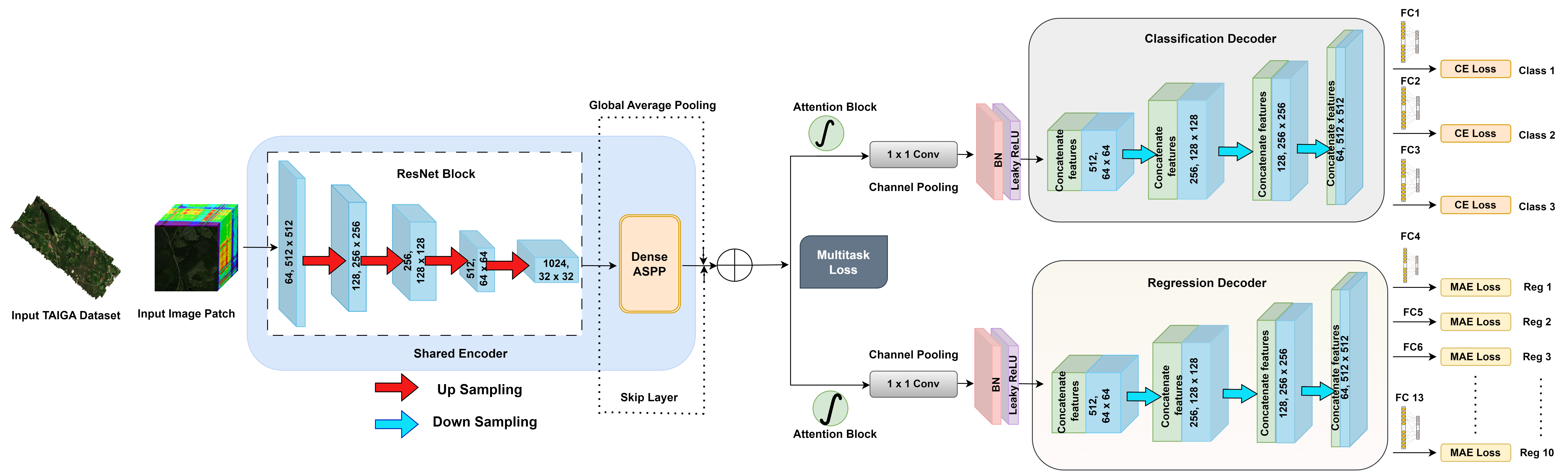}
\caption{Framework of the proposed multitask deep learning model for classification and regression of hyperspectral image. Each output has its own loss criterion, Cross-Entropy (CE) for classification, and Mean Absolute Error (MAE) for regression.}
\label{Fig1:Method}
\end{figure*}

The encoder-decoder-based U-Net model~\cite{ronneberger2015u} and its variant have been successfully developed for pixel-wise image classification and regression tasks in remote sensing. Inspired by these models, we propose an end-to-end multitask deep-learning model in this study. As shown in Figure~\ref{Fig1:Method}, the proposed method consists of three parts: a shared encoder network, a spectral-spatial attention block, and a task-specific decoder. We considered the encoder-decoder architecture to design our proposed model because of its ability to solve image-to-image prediction problems. Considering task correlations, we have considered a standard yet efficient encoder and proposed an original decoding strategy relevant to multitask learning purposes. This section will introduce the dataset considered, discuss data preprocessing and model architecture, and provide detailed insights into the multitask loss algorithms employed in our model.

\subsection{Dataset}

We consider the recently proposed Artificial Intelligence dataset for Forest Geographical Applications (TAIGA) hyperspectral dataset~\cite{9673792}. TAIGA is an airborne hyperspectral imagery acquired with the AISA Eagle II hyperspectral line scanner in the southern boreal zone of Finland. It has a spectral resolution of 4.7 nm full-width half maximum (FWHM) and a spatial resolution of 0.7 m. It consists of 13 forest variables, among which three are categorical (classification part), and the remaining ten belong to continuous data (regression part), as shown in Table~\ref{Tab1: data}. It is designed for realistic multitask-learning purposes, having continuous and categorical forest variables. Also, it is large enough to validate models at different scales and reliably identify their generalization behavior.

\begin{table}[]
\caption{Forest Variables in the TAIGA dataset}
\label{Tab1: data}
\resizebox{1.\columnwidth}{!}{%
\begin{tabular}{|c|l|c|}

\hline S.No. & Variable name & No. of classes or range \\
\hline \multicolumn{2}{|c|}{ Categorical } \\
\hline 1 & Fertility class & 4 \\
2 & Soil class & 2 \\
3 & Main tree species & 3 \\
\hline \multicolumn{2}{|c|}{ Continuous } \\
\hline 4 & Basal area $\left[\mathrm{m}^2 / \mathrm{ha}\right]$ & $0-35.51$ \\
5 & Mean DBH [cm] & $0-30.89$ \\
6 & Stem density [1/ha] & $0-6240$ \\
7 & Mean height [m] & $0-24.16$ \\
8 & Percentage of pine & $0 \%-100 \%$ \\
9 & Percentage of spruce & $0 \%-84 \%$ \\
10 & Percentage of birch & $0 \%-58 \%$ \\
11 & Woody biomass [tn/ha] & $0-180$ \\
12 & Leaf area index & $0-9.66$ \\
13 & Effective leaf area index & $0-6.45$ \\
\hline
\end{tabular}
}
\end{table}

\subsection{Data Preprocessing}
\label{dp}

\begin{figure}[!h]
 \includegraphics[width=1\columnwidth,valign=t]{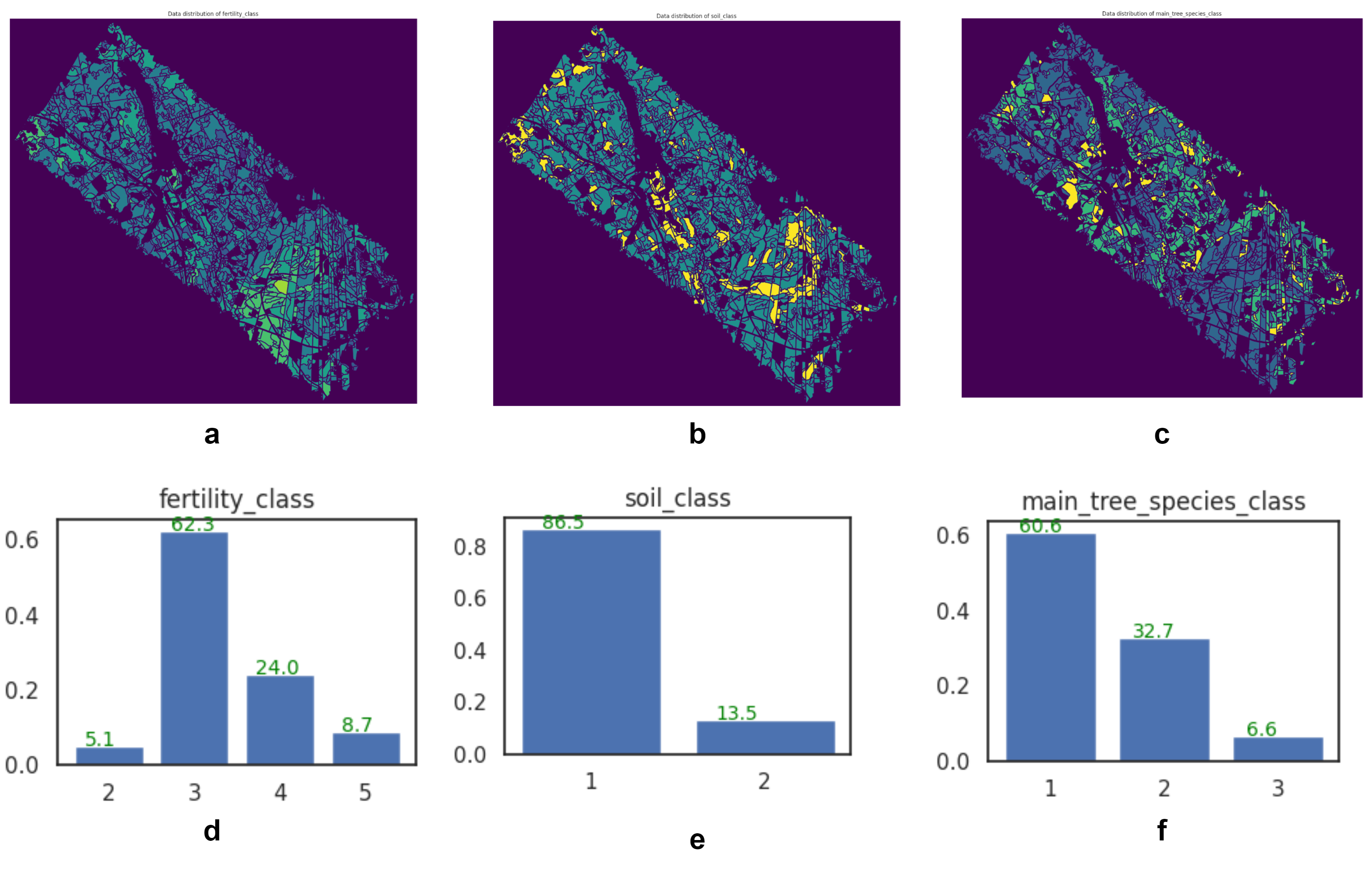}
\caption{Visualization of classification labels (a-c) and class distribution (d-f) for different categorical variables, namely, fertility class, soil type, and main tree species.}
\label{Fig2}
\end{figure}

We computed the data distribution for discrete (categorical) and continuous variables in the preprocessing section. Computing the data distribution of different classes in a large dataset is essential for understanding the inherent patterns and imbalances within the data. This information is crucial for developing effective deep learning models, as it helps to address the issues related to class imbalance and optimize training strategies. Additionally, analyzing data distribution helps in identifying potential challenges associated with imbalanced datasets, such as the presence of outliers or skewed features. This helps in designing effective data preprocessing strategies and balancing out the majority and minority classes, thus enhancing the model's ability to predict minority classes accurately. In accordance with the original TAIGA dataset paper by~\cite{9673792}, we excluded categorical and continuous classes that represent less than $0.1\%$ of the total dataset, such as the development class, diameter at breast height, etc. This helps in removing extremely skewed classes while still preserving the overall imbalanced nature of the dataset. Figure~\ref{Fig2} shows the distribution of classes for the categorical variable after removing all the extremely skewed classes. 

\begin{figure}[!h]
 \includegraphics[width=1\columnwidth,valign=t]{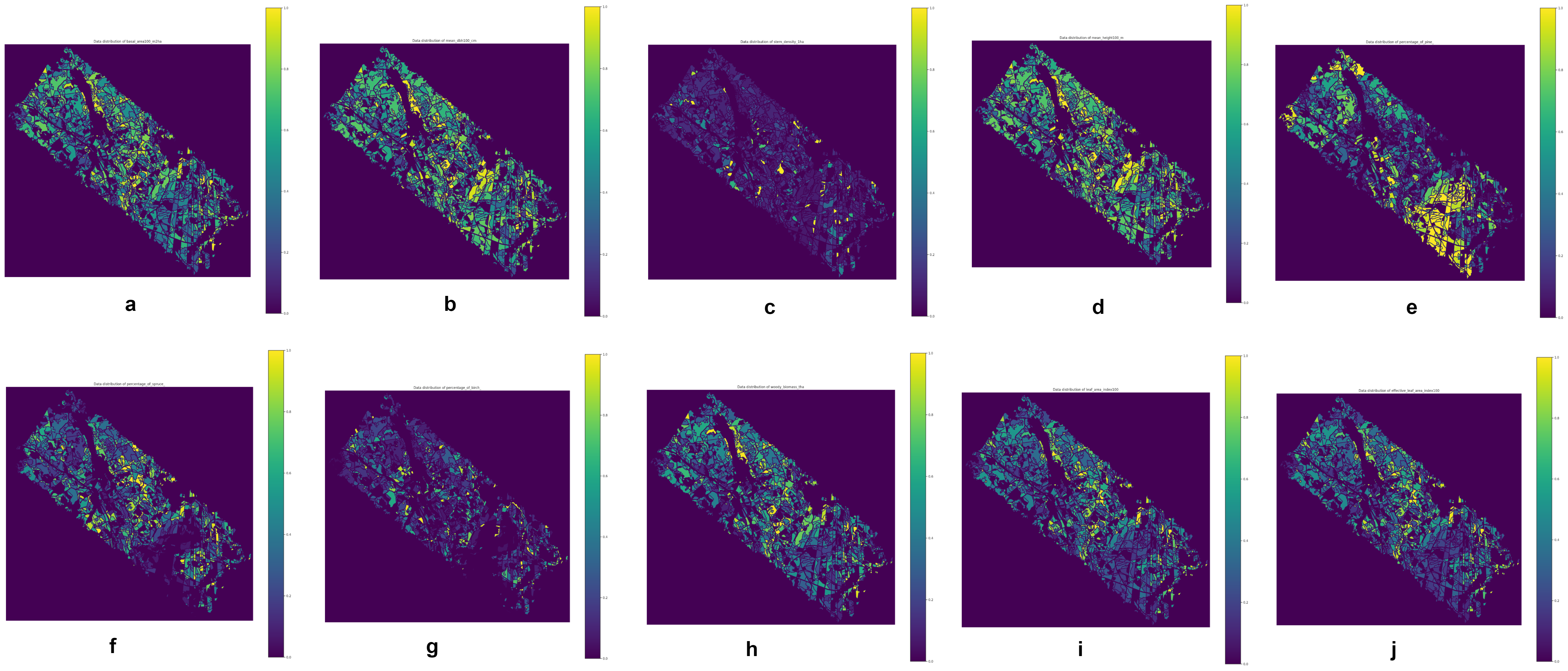}
\caption{Visualization of normalized regression labels from 0-1 for all ten continuous variables, viz., a) basal area b)mean dbh c)stem density d)mean height e)\% of pine f) \% of spruce g) \% of birch h)woody biomass i)leaf area index j) effective leaf area index}
\label{Fig3}
\end{figure}

 Further, the regression values of the continuous variables are normalized between 0 and 1 using min-max scaling since the range of values is different for each variable. This transformation brings all the continuous variables to a common scale, ensuring they share a consistent range. This approach balances the impact of variables with different scales and enhances the interpretability of regression coefficients. Figure~\ref{Fig3}  shows the labels of all ten continuous variables. 

\begin{figure}[!h]
 \includegraphics[width=1\columnwidth,valign=t]{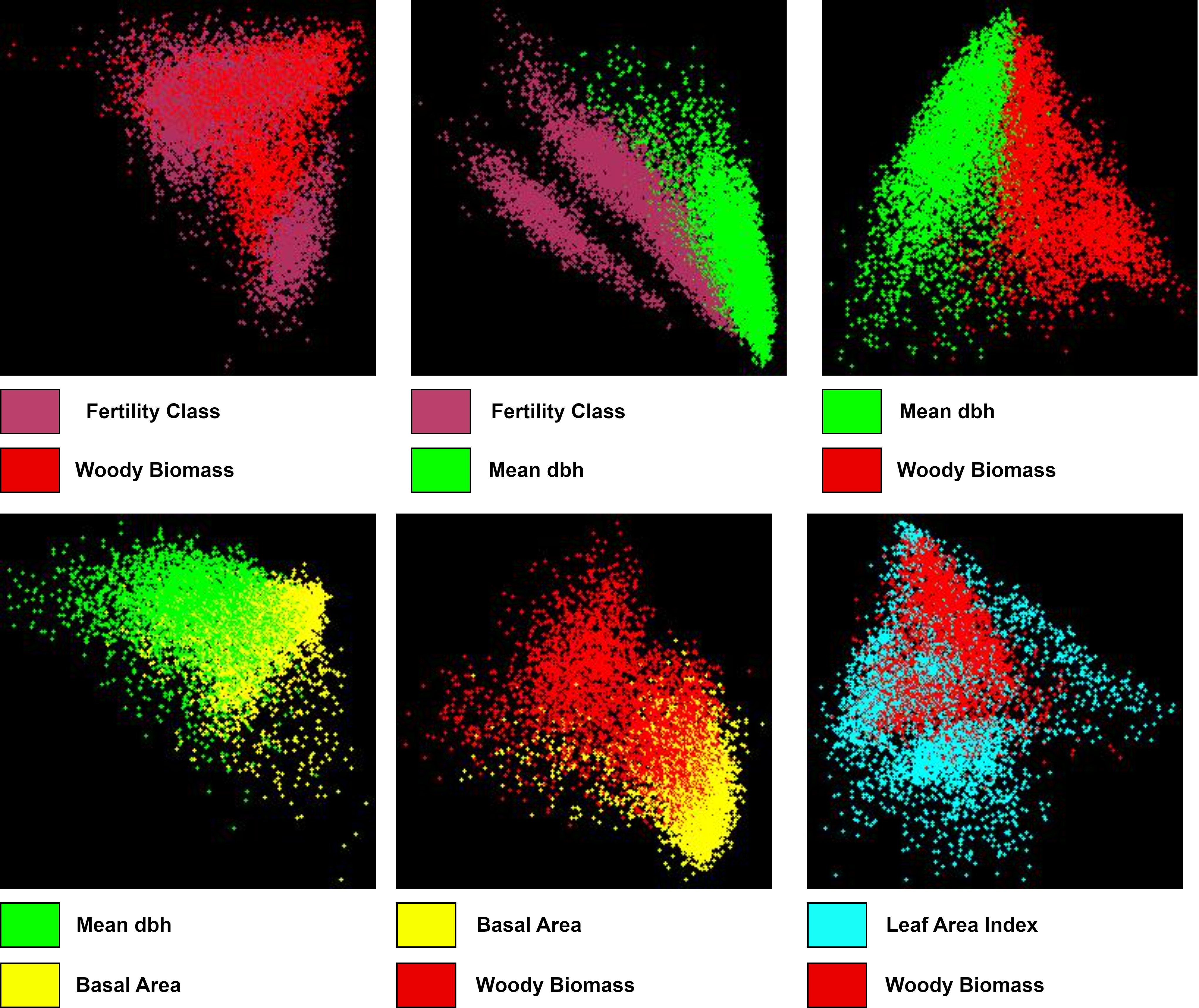}
\caption{Visualization of ROI separability to know the class overlap and correlation between different forest variables (continuous and categorical)}
\label{Fig4}
\end{figure}

Additionally, we also performed a separability analysis to understand the degree of distinction between classes. This analysis aimed to identify features that play a significant role in separating different classes from one another. We computed ROI (Region of Interest) separability using the Jeffries-Matusita (JM) distance and Transformed Divergence (TD) method (Figure~\ref{Fig4}). The JM distance quantifies the separability by transforming the multivariate data within each ROI, providing a measure of how well-defined the probability distributions of features are between the regions. Higher JM distances indicate increased separability, aiding in effective region classification. On the other hand, the TD method enhances ROI separability by transforming the original data to a space where differences between regions are prominent. This method measures dissimilarity between the probability distributions of features in different ROIs, contributing to a more discriminative analysis.

\begin{figure}[!h]
\includegraphics[width=1\columnwidth,valign=t]{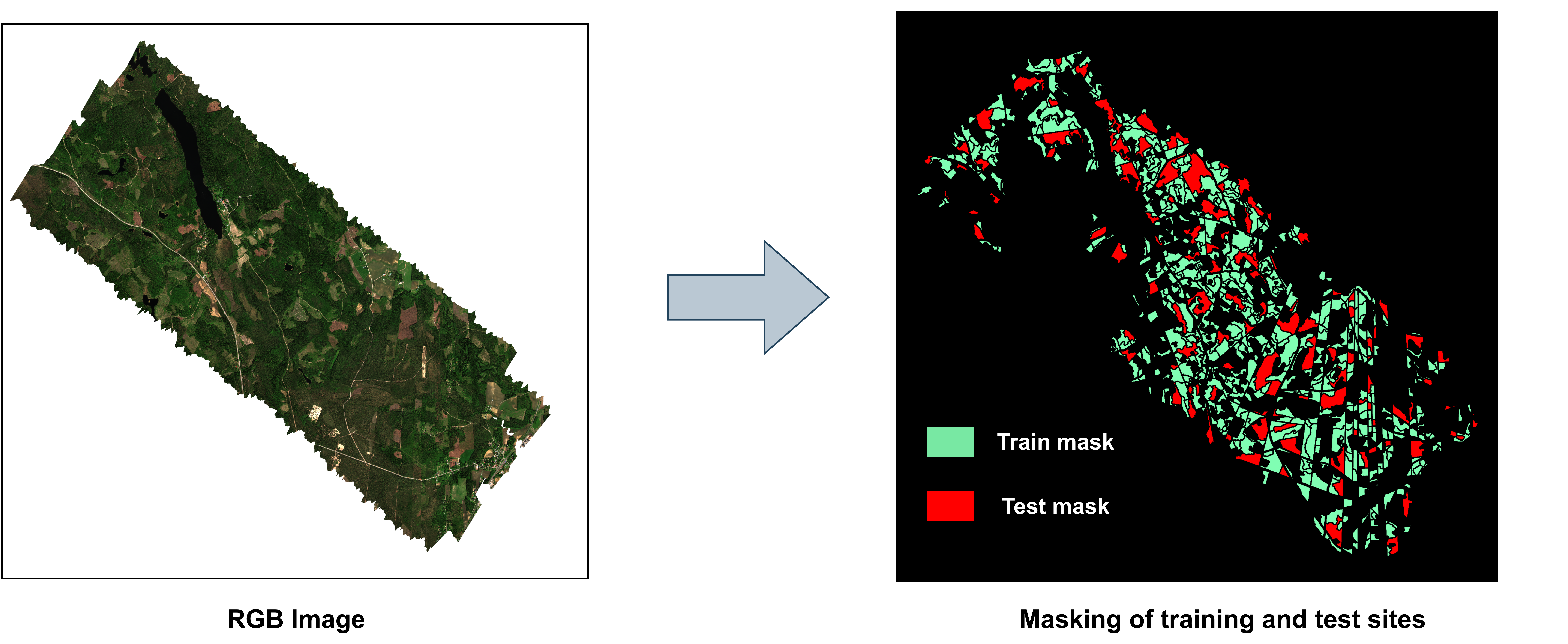}
\caption{The split dataset into training and test sets such that no two samples overlap each other and not a single pixel should present in both training and testing sets.}
\label{Fig5}
\end{figure}

The main idea is to know the correlation between different variables (continuous and categorical) since all the variables belong to the forest biome and thus have similar spatial resolution. Understanding the degree of distinction between classes is crucial for feature selection and model building to identify the most influential factors in the separation of distinct categories within the dataset. Subsequently, the dataset was split into training, validation, and test sets such that no two samples from the training and test data should overlap each other. Further, it can be observed that training and test samples are kept in separate areas such that no spatial correlation bias is introduced during the experiments (Figure~\ref{Fig5}). We replicate the data-splitting approach proposed in~\cite{9673792} to produce comparable results. 

\subsection{Shared Encoder Block}
An encoder is a crucial component in neural network architectures, transforming input data into a compressed latent representation and capturing essential features while discarding irrelevant details with respect to the objectives. The proposed shared encoder consists of a Residual Network (ResNet) block and a Dense Atrous Spatial Pyramid Pooling (ASPP), which are well-established and widely used building blocks in deep learning architectures.

\subsubsection{ResNet}
Residual Networks (ResNets)~\cite{he2016deep} is a type of deep neural network architecture that addresses the vanishing gradient problem by introducing shortcut connections, allowing the model to learn residual mappings. These shortcuts facilitate the training of very deep networks, improving performance in tasks such as image classification and regression tasks. The choice of the ResNet model for the shared encoder in our architecture is driven by its intrinsic advantages in training deep neural networks and achieving state-of-the-art performance across various tasks, including multitask learning models. ResNet consists of residual blocks that enable the training of networks with hundreds of layers, which is critical for tasks requiring complex feature extraction and classification. Compared to alternative architectures such as VGG (Visual Geometry Group), Inception, DenseNet, or MobileNet, ResNet's combination of depth, performance, transfer learning capabilities, and extensive empirical validation makes it an optimal choice for our architecture's shared encoder. The availability of resources, tools, and optimizations further supports our decision to utilize ResNet to enhance model efficacy and efficiency.

\begin{figure}[h!]
\centering
\subfloat[]{\includegraphics[width=0.5\columnwidth]{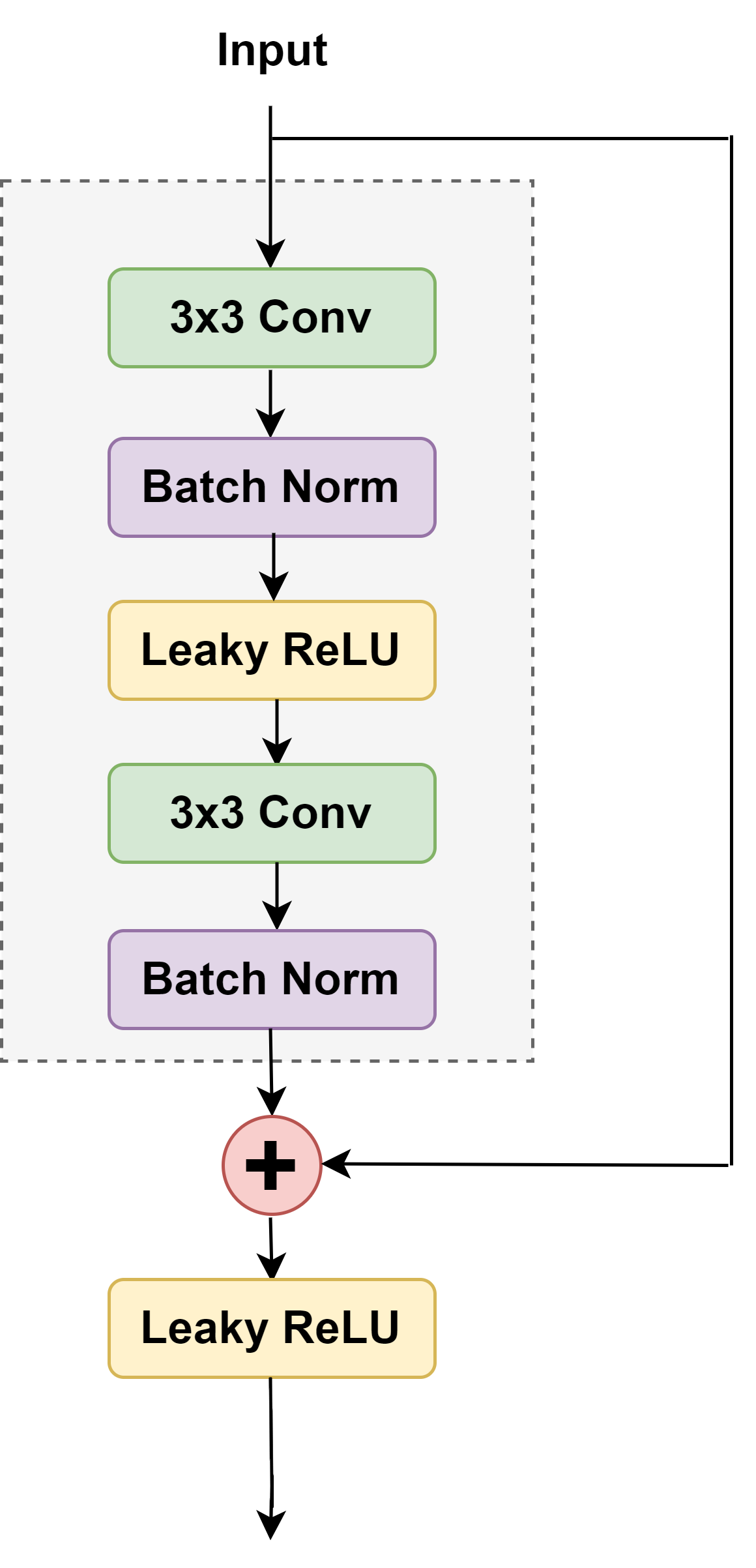}}\hfill
\subfloat[]{\includegraphics[width=0.65\columnwidth]{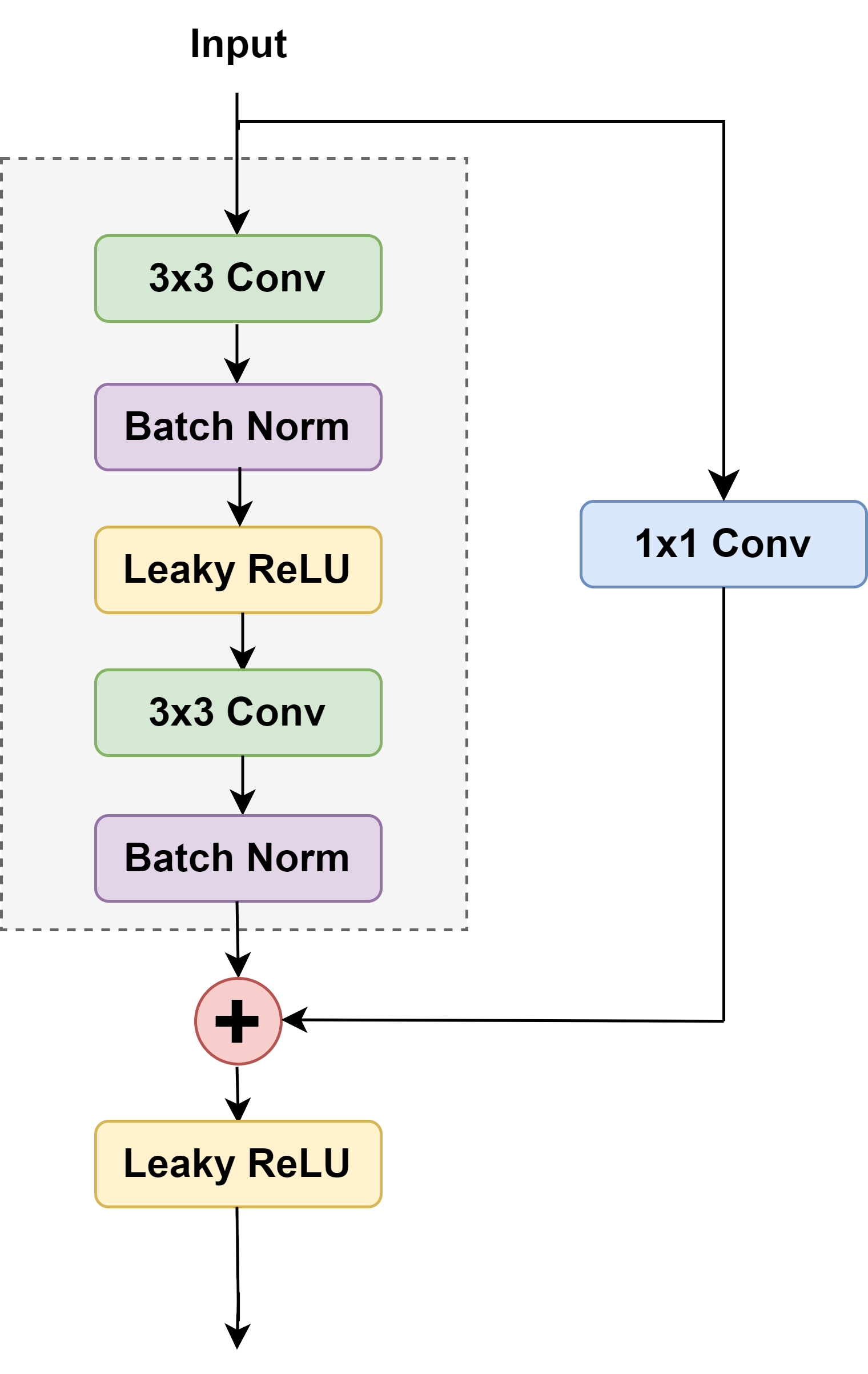}}\hfill

\caption{ResNet Block. a) Identity ResNet Block  b)  ResNet block with a 1x1 convolution that transforms the input into the desired shape for the addition operation called "bottleneck block."}
\label{Fig6}
\end{figure}

As shown in Figure~\ref{Fig6}, the encoder network begins by employing a convolution block to augment the number of feature dimensions. This block comprises a 2-D convolution layer with a $3\times3$ kernel size, followed by a batch normalization layer and an activation layer. The choice of a smaller kernel size aims to prevent information loss from input images by summarizing features across pixels. Batch normalization is applied to stabilize and expedite training, while the activation layer introduces nonlinearity to the model. Leaky ReLU is selected as the activation function instead of the commonly used ReLU. This decision is motivated by the observation that neurons cease learning when the ReLU function enters the negative range. Leaky ReLU addresses this limitation by allowing a small negative slope for negative inputs.

Following this is the cascade of the residual network. Let R(x) represent the final output of the residual network; then,
\begin{equation}
    F\left ( x \right ) = R\left ( x \right ) - x
\end{equation}
 where x denotes the input image and $F\left ( x \right )$ represents the residual mapping (the dotted line in figure~\ref{Fig6}). The above-mentioned equation is called a shortcut connection in a feedforward network and can also be represented as
\begin{equation}
    y_{i}= F\left ( x_{i} , \left \{ w_{i} \right \}\right ) + x_{i}
\end{equation}
where $y_{i}$ and $x_{i}$ represent the output and input vectors of the $i^{th}$ set of stacked layers, and $F\left ( x_{i} , \left \{ w_{i} \right \}\right )$ is the shortcut connection. For consistency with the input tensor $x$, the shortcuts in our model must have equal sizes. These shortcuts can be mapped as an identity or through a linear projection to match dimensions. Figure~\ref{Fig6}b showcases a ResNet block, a key component in our network construction. Within the ResNet block, each convolutional layer is followed by a batch normalization. Our proposed model adjusts the number of channels in $x_{i}$ within the ResNet block. Consequently, our model employs the ResNet block with a projection shortcut with $1\times1$ conv layer as it can learn more complex features specifically on larger datasets and have better scalability.

\subsubsection{Dense ASPP}
A multiscale approach is required since the accurate analysis of complex tasks, such as simultaneous classification and regression, demands robust feature representation across various spatial resolutions. In our proposed framework, the integration of Dense ASPP~\cite{chen2018encoder} is instrumental in achieving this goal effectively. DenseASPP is an extension of the Atrous Spatial Pyramid Pooling (ASPP) module, which captures multi-scale contextual information. ASPP employs atrous (dilated) convolutions at multiple dilation rates, thereby enabling the network to perceive both local and global context information simultaneously. Initially developed for semantic segmentation, ASPP has proven its versatility in various network applications by efficiently extracting multi-scale features to attain a substantial receptive field. In our study, we advanced the ASPP module by incorporating Dense connectivity, resulting in the Dense ASPP modification. Its dense connectivity enhances feature reuse across tasks, leading to improved model efficiency and performance in jointly learning classification and regression tasks, all achieved without a significant increase in computational cost.

The atrous convolution can be formulated as follows:
\begin{equation}
    o[i] = \sum_{k=1}^{F} y[i + r.k]. w[k]
\end{equation}

\noindent where $o$ and $y$ are the output and input signals (output of ResNet), respectively. $F$ is the filter size, $w[k]$ represents the $k^th$ parameter of the filter, and r is the dilation rate. Each atrous layer's result is combined with the input feature map and the outputs from preceding lower layers. This concatenated feature map is then forwarded to the subsequent layer in the network. The ultimate output of Dense ASPP comprises a feature map produced through the utilization of multi-rate, multi-scale atrous convolutions. This innovative architecture enables the simultaneous construction of a more comprehensive and intricate feature pyramid with minimal dependence on the extensive use of atrous convolutional layers. The Dense ASPP module is typically expressed as a concatenation of dilated convolutions with different dilation rates. The mathematical expression for a Dense ASPP module with dilation rates $r_{1}$, $r_{2}$,......$r_{n}$ can be written as:
\begin{equation}
Y= Concat(Conv_{r_{1}}(X,W_{r_{1}}),.....Conv_{r_{n}}(X,W_{r_{n}}))
\end{equation}

\begin{figure}[!h]
 \includegraphics[width=1\columnwidth,valign=t]{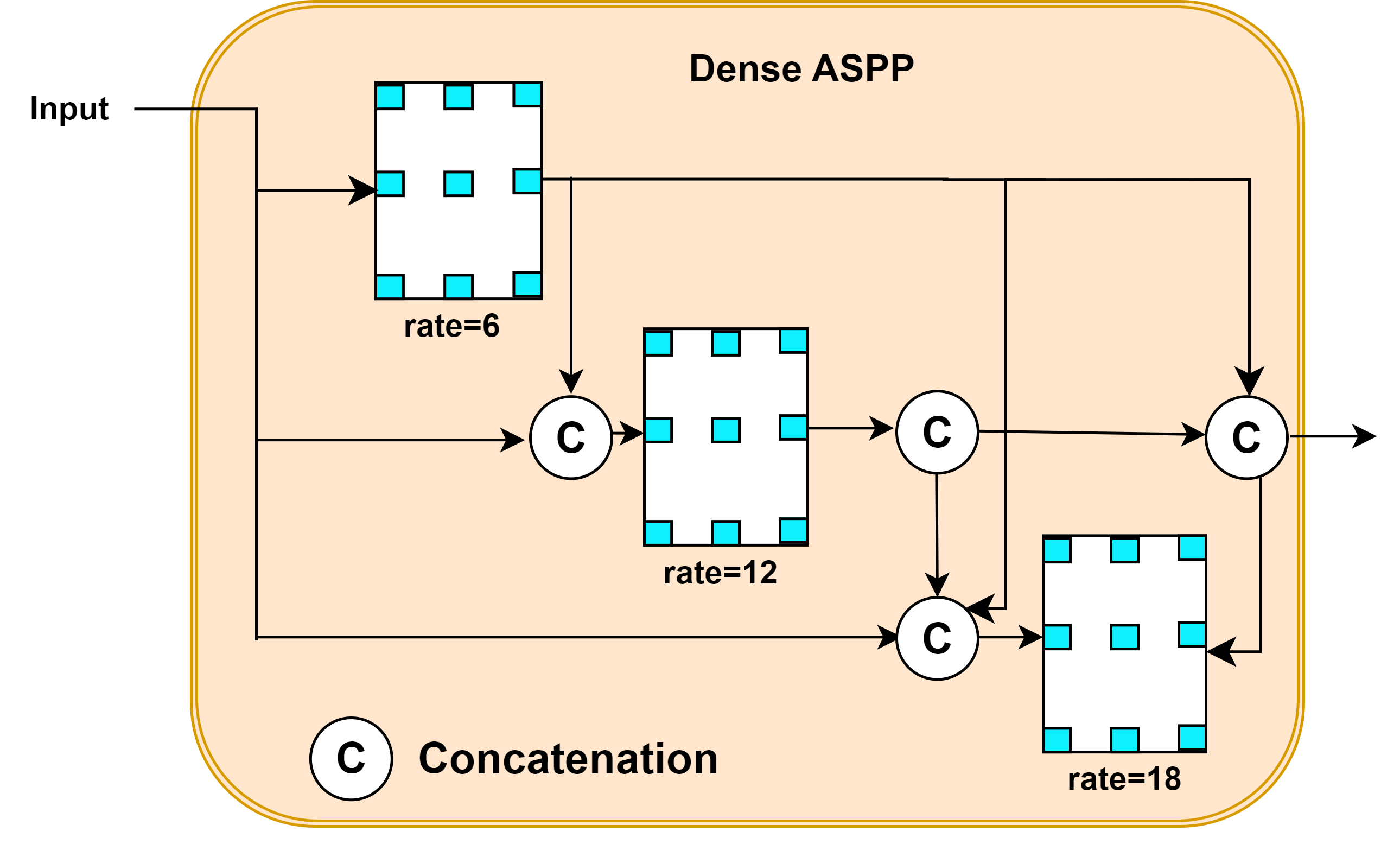}
\caption{The architecture of Dense ASPP with dilation rate of 6, 12, and 18.}
\label{Fig7}
\end{figure}

\noindent Here $Conv_{r_{1}}(X,W_{r_{1}})$ denotes the dilated convolution with dilation rate $r_{n}$ applied to the input feature map 
$X$ using the corresponding set of filters $W_{r_{i}}$. The $Concat(…)$ operation combines the feature maps obtained from these dilated convolutions along the channel dimension. The dilation rate considered for the proposed work is $6$, $12$, $18$ (Figure~\ref{Fig7}).

DenseASPP, through its integration of ASPP and dense connectivity, aims to improve the performance of the network by integrating contextual information, making it suitable for tasks like classification and regression. The ASPP module enhances the model's ability to capture contextual information, proving valuable for classification and regression by considering broader understanding. Meanwhile, dense connectivity ensures efficient training by addressing issues like the vanishing gradient problem. Furthermore, the flexibility in architecture design meets the specific requirements of different tasks within a multitask learning framework. We adopt summation instead of concatenation because summation reduces feature dimensions and the number of parameters in the next layers thus simplifying model and making training easier. The integration of multiple reception information from ResNet and Dense ASPP can be formulated as follows:

\begin{dmath}
  F_{l}= f_{l-1} + \sum_{i=1}^{N} \frac{1}{1+e^{-\lambda _{i}}} \times BN \left (w_{1\times 1, 0}^{c}\left ( A\left ( BN\left ( w_{3\times 3, _{d_{i}}}^{1.5.h}\left(f _{l-1} \right ) \right ) \right ) \right )  \right )  
\end{dmath}

where $f_{l-1}$ denotes the input features, and c denotes the number of bands of the input features. BN(·) and A(·) represent the batch normalization operation and Leaky ReLU activation function, respectively. $w_{3\times 3, _{d_{i}}}^{1.5.h}$ denotes $3\times3$ convolutional kernels with $d_{i}$ as the dilation rate and $1.5.h$ as the number of bands of the output features. $w_{1\times 1, 0}^{c}$ indicates $1\times1$ convolutional kernels with no dilation, and the number of bands of the output features is equal to $h$. The output of the summation layer is subjected to a global average pooling layer to halve the original size in the spatial dimensions, followed by a skip layer to increase the feature channels to twice the amount of the input channels, which can be formulated as follows:

\begin{equation}
    f_{l}= A\left ( BN\left ( w_{1\times 1, 0}^{2 \cdot  c} \left ( GAP\left ( F_{l} \right ) \right )\right ) \right )
\end{equation}

where GAP(·) and $f_{l}$ denote the average pooling layer and the output of the summation, respectively. $w_{1\times 1, 0}^{2 \cdot  c}$ denotes
$1\times1$ convolutional kernels with no dilation, and the number of bands of the output features equals $2 \cdot  c$.

\subsection{Spectral-Spatial Residual Attention Block}
\begin{figure*}[!h]
 \includegraphics[width=\textwidth]{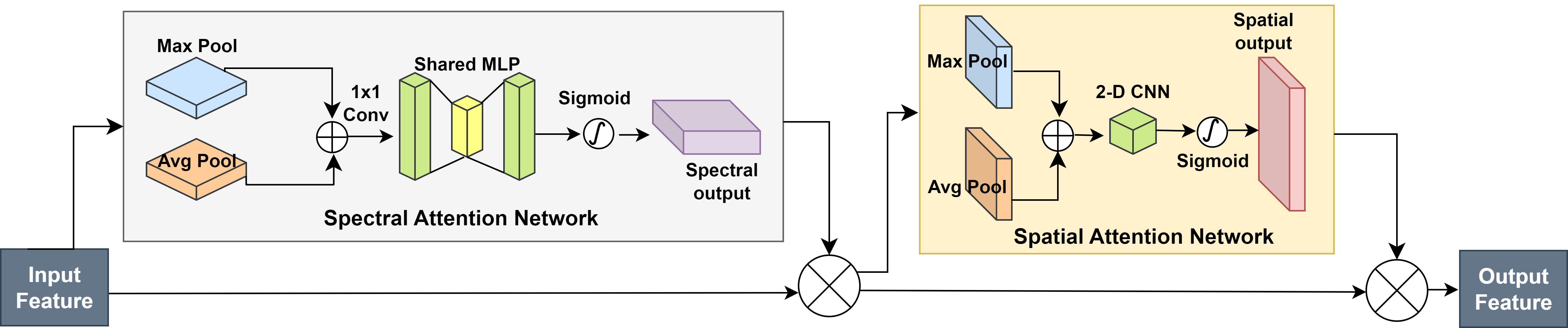}
\caption{Framework of spectral-spatial residual attention block based on convolutional block attention module}
\label{Fig8}
\end{figure*}

The attention network in deep learning models is introduced to focus more on important features and discard unnecessary features~\cite{vaswani2017attention}. By enabling selective information processing, these modules allow the model to dynamically focus on relevant parts of the input for each task, enhancing its adaptability to different requirements. The ability to selectively attend to task-specific features improves generalization, making the model more robust to unseen data and tasks. In a typical attention network, a weight matrix is generated from the input and is applied back to itself. Woo et al.~\cite{woo2018cbam} introduced CBAM (Convolutional Block Attention Module), a simple yet effective attention network for CNNs. An attention network can be formulated as follows:
\begin{equation} 
R_{l} = R_{l-1} . A(R_{l-1})
 \end{equation}

\noindent where, $R_{l}$ and $R_{l-1}$ represent input and output of the residual unit of $l_{th}$ layer and the function $A(.)$ represents operations such as pooling, fully connected, ReLU, and sigmoid function. In this work, we have proposed the combination of attention and residual network called residual attention network to enhance feature representation and improve gradient flow, which can be represented as:
\begin{equation} 
R_{l} = R_{l-1} + F(R_{l-1}) . A(F(R_{l-1}))
 \end{equation}

where F represents the feature map. As shown in the figure~\ref{Fig8}, the sequential feeding of input features into spectral attention modules and spatial attention modules is a strategic approach aimed at empowering learning-based networks to discover both "what" and "where" to focus during information processing. A channel attention map is generated by first passing the input through a spectral attention module, allowing the network to grasp "what" to prioritize in the channel dimension. This mechanism enhances the model's ability to selectively emphasize relevant features within each band, providing a foundation for capturing critical patterns and information. The subsequent utilization of spatial attention modules further refines this process, enabling the network to understand "where" in the spatial domain to concentrate its attention. 

Within the channel attention module, the input features undergo a dual transformation through a max-pooling layer and an average-pooling layer, resulting in two distinct vectors. These pooling feature maps are combined and fed as input to the attention network and are calculated as follows:

\begin{equation}
Z= P_{avg} + P_{max}
\end{equation}

where, \begin{equation}
P_{avg} = Avg (F)
        = \frac{1}{h\times w}\sum_{i=1}^{h}\sum_{j=1}^{w} F_{i,j}
\end{equation}
\begin{equation}
P_{max} = max (F)
\end{equation}

A feature map $F$ $\epsilon$ $ \mathbb{R}^{h\times w\times d}$, where $h\times w$ denotes spatial size and $d$ denotes number of spectral bands, and $F_{i,j}$ denotes the value at position ($i,j$) of the input $F$. A 1-D CNN is used for mapping spectral vectors and extracting spectral features, followed by a shared multilayer perception (MLP). Two bottleneck fully connected MLP layers are considered to enhance the extracted spectral features. This layer tends to reduce model complexity and support model generalization ability. The first layer, represented as $W_{1}$, is a dimensionality reduction layer, and the second layer, represented as  $W_{2}$, is a dimensionality increment layer. The spectral features are calculated and can be formulated as follows:
\begin{dmath}
F_{spectral} = \delta \left ( W_{2}\left ( W_{1}\left ( P_{Avg} \right ) \right ) \right ) + W_{2}\left ( ReLU\left ( W_{1}\left ( P_{Max} \right ) \right ) \right)
\end{dmath}

\noindent where $\delta$ represents the Sigmoid function and ReLU is the hidden activation of the shared MLP.
Besides focusing on "what" to prioritize, the results undergo further refinement in a spatial attention module, generating spatial attention maps that help the network on "where" to focus. Spatial attention maps are computed using max-pooling and average-pooling along the channel axis to generate two feature maps. These feature maps are concatenated and processed through a $3\times3$ convolution layer, followed by a sigmoid function to limit attention values between 0 and 1. The final step involves a multiplication operation between the input features and the 2-D spatial attention map along the spatial dimensions. The rich low/mid-level features are transmitted to the decoder portion through attention modules in the channel pooling layer. This is performed before the first upsampling step, where upscaled features are compressed separately for regression and classification tasks to 256 dimensions with a $1\times1$ kernel. As shown in section~\ref{dp}, classification and regression tasks indeed need decorrelation that facilitates the amplification of meaningful information while suppressing irrelevant features. The channel pooling layer captures complex patterns or relationships between different channels in the feature map by adjusting the weights in the $1\times1$ convolutions with adaptive channel interactions. 

\subsection{Task-Specific Decoder}
In the multitask learning model, task-specific decoders allow the individual adjustment of parameters, accommodating variations in task complexity and requirements. This capability to independently customize adjustments aids in isolating task-specific features throughout the training process, thereby effectively reducing interference between tasks. We apply batch normalization to each layer to concatenate the resulting features together. This produces the shared representation between each task. A Leaky ReLU activation function follows this to improve the nonlinear learning ability of the model. Once all shared features are extracted, the multitask decoder is designed, and we first separate the classification and regression tasks into two distinct paths, each being more task-specific.

We deliberately kept every decoder block symmetrical to the shared encoder block, and the features of the corresponding encoding layer concatenated with the task-specific decoder (Figure~\ref{Fig1:Method}). This is to maintain the multiscale features, enhance important channel features, and weaken unimportant channel features. As mentioned in section~\ref{dp}, computing class overlap and feature correlation helps in designing dedicated ResNet-based decoders for each path by identifying and utilizing common features that are beneficial for both classification and regression tasks. This approach allows each decoder to specialize in refining these shared features according to the specific needs of each task by reducing task interference.
Finally, task-specific heads are optimized, relying on a single fully connected layer. Bilinear upsampling is applied to scale the output to the same resolution as the input. Most of the model’s parameters and depth are thus in the feature encoding, with very little complexity in each task decoder.  This highlights the advantage of multitask learning, where computational resources can be efficiently shared to learn a more effective shared representation. By employing task-specific loss functions, such as cross-entropy (CE) for classification and mean squared error  (MAE) for regression, the model optimizes parameters independently for each task, enhancing training efficiency. 


\subsection{Multitask Loss}
Loss functions are crucial in determining how models assess the overall discrepancy between predicted outcomes and ground truth. The selection of a loss function has a direct impact on both the learning process and the ultimate outcomes of models. Even a well-designed model can be affected if paired with an inappropriate loss function. Further, when considering the multitask model, loss balancing must be studied. The multitask loss function combines the individual losses associated with each task into a single composite loss, which the model aims to minimize during training. The formulation of a multitask loss typically involves assigning weights to each task's contribution to the overall loss, reflecting the relative importance of different tasks. This allows the model to strike a balance between optimizing for various multitask objectives such as classification and regression. In this research, the multitask loss function has been used for different purposes: 1) to handle class imbalance and 2) task balancing for multitask learning. 

In the given dataset, the class distribution is not uniform, i.e., there are a few minor classes that are less than 10\% of the whole dataset. This leads to data imbalance, resulting in the consideration of major classes and the ignoring of minor classes. In dealing with class imbalance, it is crucial to appropriately adjust learning goals and metrics. Attempts to address class imbalance often involve sampling strategies, such as oversampling minority classes or undersampling majority classes. However, these approaches have limitations. Oversampling can lead to overfitting, while undersampling may discard valuable information. In multi-label, multi-class scenarios, the complexity increases as a sample may belong to a minority class for one label and a majority class for another, making traditional resampling less effective. Thus, in this work, to handle the class imbalance in the dataset, instead of using a sampling method, two different methods have been adopted, naming cost-sensitive learning ~\cite{qin2010cost} and focal loss ~\cite{lin2017focal} (without concurrent methods). 

Cost-sensitive learning considers the misclassification cost, thus reducing the performance accuracy in imbalanced datasets where the algorithm is biased towards majority classes. Let $c_{i}$ be the cost of classifying a sample for class $i$ and $p_{i}$ be the frequency of class $i$ in the dataset. The cost for the wrong prediction for class $i$ is computed as ~\cite{ling2008cost}: 
\begin{equation}
c_{i} = \frac{1}{p_{i}}
\end{equation}

\noindent However, this doesn't work for an extremely imbalanced dataset. For our multitask imbalanced dataset, we considered inverse median frequency for computing classification cost. Let $p_{m}$ be the median of class frequencies, then the cost for the wrong prediction for class $i$ is computed as: 
\begin{equation}
c_{i} = \frac{p_{m}}{p_{i}}
\end{equation}
 
\noindent The standard cross-entropy loss assumes all classification errors are equally important, which fails for class-imbalanced datasets. Thus, the focal loss dynamically scales the standard cross-entropy loss. The focal loss ~\cite{lin2017focal} for a multitask imbalanced dataset can be computed as:
\begin{equation}
\iota _{f}= -\alpha \left ( 1-p \right)^{\gamma }\log p
\end{equation} 
\noindent where $\alpha$ is the balanced factor and $\gamma$ is the focusing parameter. With reference to the work by~\cite{pham2019deep}, we considered the value of $\alpha$ as $0.25$ and $\gamma$ as $2$. $p\epsilon \left [ 0,1 \right ]$ is the predicted value of the given data sample. 

Additionally, training the multitask model requires optimizing each loss function weight to improve the overall performance accuracy. We have considered two different approaches to address this issue: uncertainty loss and gradient normalization (GradNorm) loss. 

\subsubsection{Uncertainty Loss}
The uncertainty loss of every single task in classification and regression can be computed using Kendall et al. ~\cite{kendall2018multi} assumption for the multitask loss function. This method, known as Homoscedastic uncertainty loss, refers to a type of uncertainty that remains constant regardless of input data. Instead, it varies between tasks and is task-dependent. In multi-task learning, homoscedastic uncertainty can effectively capture the relationship between related tasks and thus aim to fine-tune task weightings through probabilistic modeling.

Considering $f\left ( x; W \right )$ as the output of the deep neural network, where $W$ is the weight and $x$ is the input. Let Gaussian distribution be the considered likelihood for the regression model with mean $f\left ( x; W \right )$ and variance $\sigma ^{2}$ given as:
\begin{equation}
    p\left ( y|f\left ( x; W \right ) \right )= N\left ( f\left ( x; W \right ), \sigma ^{2} \right )
\end{equation}

The output likelihood is typically passed through a softmax function for classification tasks where Kendall et al. utilized a scaled version of the model output given as:

\begin{equation}
    p\left ( y|f ( x; W \right ), \sigma ) = Softmax \left (  \frac{1}{\sigma ^{2}}f\left ( x; W \right )\right )
\end{equation}

\noindent where $\sigma$ is a scalar noise. The model is optimized by maximizing the log-likelihood of its output. For regression tasks, the log-likelihood of the output is expressed as follows:
\begin{equation}
    \log p\left ( y|f\left ( x; W \right ) \right ) \propto -\frac{1}{2\sigma ^{2}}\left \| y-f\left ( x; W \right ) \right \|^{2}- \log \sigma 
\end{equation}

\noindent and for classification, the output is:
\begin{multline}
    \log p\left ( y|f\left ( x; W \right ), \sigma  \right ) = \log \mathrm{Softmax} \left ( \frac{1}{\sigma ^{2}} f\left ( x; W \right )\right) \\
    = \log \frac{\exp\left ( \frac{1}{\sigma ^{2}}f_{c}\left ( x; W \right ) \right )}{\sum_{c^{'}}\exp\left ( \frac{1}{\sigma ^{2}} f_{c^{'}}\left ( x; W \right )\right )} \\
    = \frac{1}{\sigma ^{2}}f_{c}\left ( x; W \right )- \log\sum_{c^{'}}\exp\left ( \frac{1}{\sigma ^{2}}f_{c^{'}}\left ( x; W \right ) \right )
\end{multline}

where $f_{c}\left ( x; W \right )$ is the $c^{th}$ element of vector $f\left ( x; W \right )$ 

Let the output for regression be $y_{1}$ and for classification be $y_{2}$. The multitask loss can be given jointly as:

\begin{align}
    \pounds \left ( W, \sigma _{1}, \sigma _{2} \right ) &= - \log p (y_{1}, y_{2} = c|f(x; W)) \notag \\
    &= - \log N (y_{1}; f(x; W), \sigma_{1}^{2}) - \log p \notag \\ 
    &\quad (y_{2} = c|f(x; W), \sigma_{2}) \notag \\
    &= \frac{1}{2 \sigma_{1}^{2}} \left \| y_{1} - f_{c} (x; W) \right \|^2 + \log \sigma_1 \notag \\
    &- \frac{1}{\sigma_{2}^{2}} f_{c}(x; W) \notag \\
    &\quad + \log \sum_{c^{'}} \exp \left( \frac{1}{\sigma_{2}^{2}} f_{c^{'}}(x; W) \right) \notag \\
    &= \frac{1}{2 \sigma_{1}^{2}} \pounds_1 (W) + \log \sigma_1 - \frac{1}{\sigma_{2}^{2}} f_c (x; W) \notag \\
    &\quad + \log \sum_{c^{'}} \exp \left( f_c{'}(x; W) \right)^{\frac{1}{\sigma_{2}^{2}}} \notag \\
    &\quad + \log \sum_{c^{'}} \exp \left( \frac{1}{\sigma_{2}^{2}} f_{c^{'}}(x; W) \right) \notag \\
    &\quad - \log \sum_{c^{'}} \exp \left( f_{c{'}}(x; W) \right)^{\frac{1}{\sigma_{2}^{2}}} \notag \\
    &\approx \frac{1}{2\sigma_{1}^{2}} \pounds_{1} (W) + \log \sigma_{1} + \frac{1}{\sigma_{2}^{2}} \pounds_{2} (W) + \log \sigma_{2}
\end{align}

\noindent
where, 
\begin{equation*}
    \pounds _{1} (W) = \left \| y_{1}- f_{c}(x; W) \right \|^{2}
\end{equation*}
and 
\begin{equation*}
    \pounds _{2}(W)= - \log \mathrm{Softmax} (y_{2}, f_{c}(x; W )).
\end{equation*}

\noindent
The final equation can be derived assuming 
\[
\frac{1}{\sigma_{2}^{2}}\sum_{c^{'}} \exp \left( \frac{1}{\sigma_{2}^{2}}f_{c^{'}}(x; W)\right) \approx \sum_{c^{'}} \exp \left( f_{c^{'}}(x; W)\right)^{\frac{1}{\sigma_{2}^{2}}}
\]
when $ \sigma _{2} \rightarrow 1 $.\\

\noindent The objective function shows that \( \sigma_1 \) and \( \sigma_2 \) control how much influence \( \pounds_1(W) \) and \( \pounds_2(W) \) have on the overall loss. This approach can be straightforwardly extended to handle multiple regression and classification outputs. In terms of implementation, the numerical stability is managed by training the logarithm of the variance given as:
\begin{equation}
    s=log\left ( \sigma ^{2} \right )
\end{equation}

\noindent where $s$ is the uncertainty loss and $\sigma$ is the log variance for likelihood of the output. 

\subsubsection{Gradient Normalization (GradNorm) Loss}
GradNorm loss is another loss-based regularization function that handles task balancing in the multitask learning model. It automatically balances the multitask loss function by directly tuning the gradients to equalize task training rates. The GradNorm is computed as ~\cite{chen2018gradnorm}:
\begin{equation}
    \iota _{grad}\left ( t; w_{i} \left ( t \right )\right )= \sum_{i}\left \| G_{w}^{i}(t)- \bar{G} w(t).[r_{i}(t)]^{\alpha} \right \|
\end{equation}

\noindent where $G_{w}^{i}(t)$ is the $L2$ norm of the gradient for a task $i$ of a weighted task loss $w$ at time step $t$. $\bar{G} w(t)$ is the average gradient norm, $r_{i}(t)$ is a relative inverse training rate, and $\alpha$ is a hyperparameter that modulates the strength of training rate balancing. The pseudo-code of the GradNorm algorithm is shown in Algorithm 1. GradNorm proved to be an efficient algorithm for multi-task learning.

\begin{figure}
    \centering
    \includegraphics[width=1\linewidth]{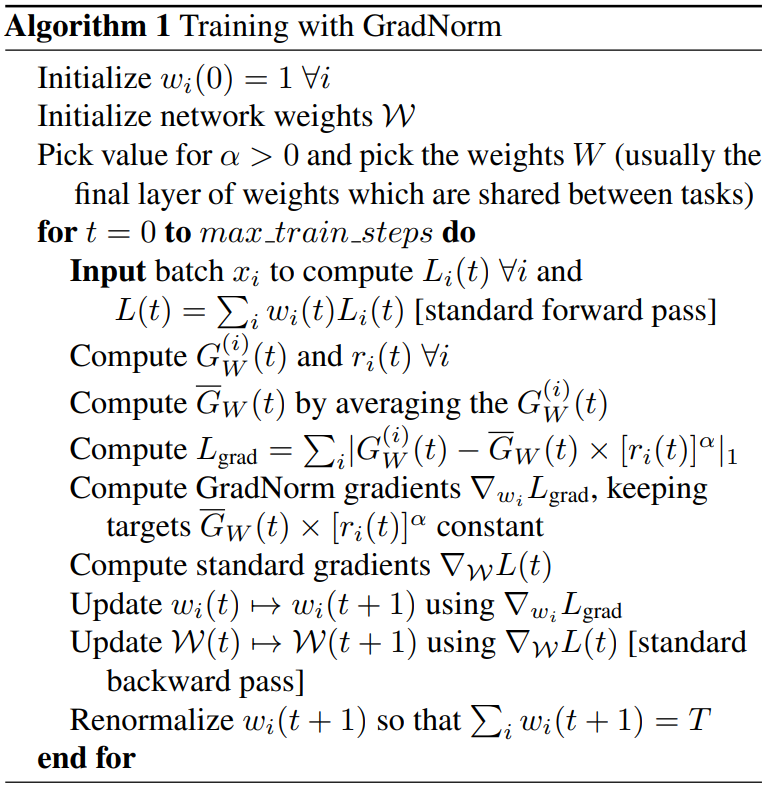}
    \label{fig:enter-label}
\end{figure}

\section{Experiments and Result Analysis}
\subsection{ Experimental Configuration and Parameter Settings}
\noindent In this paper, the experiment has been conducted on a computer with an Intel(R) Xeon(R) Silver 4214R CPU at 2.40 GHz with 64 GB RAM and an NVIDIA GeForce RTX A6000 graphical processing unit (GPU) with 51 GB RAM. The software environment is the Ubuntu 14.04 ultimate 64-bit system with a deep learning framework of PyTorch. Deep learning relies on data and is prone to overfitting. To avoid overfitting, we implemented data augmentation (random horizontal and vertical flips), ensuring that each batch of images remained diverse. Following augmentation, we standardized the images by normalizing them using min-max scaling for precalculated mean and variance. We generated prediction maps for the test dataset during the inference stage without employing data augmentation. We initialized the learning rate at $0.001$ and the Adam optimizer as a stochastic gradient descent method with a momentum of 0.7 and weight decay ($\eta$ ) of $10^-4$. We employed a learning rate warm-up strategy for the initial five epochs to enhance model stability during training. This involved setting the learning rate in the first epoch to one-fifth of the initial value and incrementing it by $0.0002$ per subsequent epoch until reaching a learning rate of $0.001$. Full-resolution images have been used to train the network for 150 epochs with a mini-batch of size 64. 

In networks designed for multitask learning, establishing an appropriate balance among various classes is crucial when training multiple tasks. To assess the effectiveness of our approach, we conducted training using the proposed method with pre-trained parameters and a frozen encoder network. We used cross-entropy loss for the classification tasks and mean absolute error loss as the loss function for the regression tasks. This is to measure the gap between the ground truth and the prediction map and to update the network parameters through backpropagation. We opted for cross-entropy loss in the context of classification tasks at the pixel level, as it is well-suited for scenarios where the goal is to measure the difference between predicted probability distributions and actual class distributions. The cross-entropy can be represented as:
\begin{equation}
\text{Cross-Entropy} = -\frac{1}{N} \sum_{j=1}^{N} \sum_{i} y_{ji} \cdot \log(\hat{y}_{ji})
\end{equation}

where $N$ is the number of samples, $y_{ji}$ represents the true label and $\hat{y}_{ji}$ represents the predicted probability for the $i^{th}$ class of the $j^{th}$ data point.
On the other hand, mean absolute error (MAE) loss was chosen for regression tasks as it measures the average magnitude of errors between predicted values and actual values. In regression problems, the goal is to predict continuous numerical values, and MAE is a suitable choice for quantifying the accuracy of predictions by emphasizing the magnitude of errors. Additionally, given that the data is scaled to small values, MAE becomes more effective in addressing small errors. The mean absolute error can be represented as:

\begin{equation}
    \text{MAE} = \frac{1}{n} \sum_{i=1}^{n} |y_i - \hat{y}_i|
\end{equation}

where n is the total number of data points, $y_i$ is the true value and $\hat{y}_i$  is the predicted value for the $i^{th}$ data point.

\subsection{Experimental Results}
\subsubsection{Quantitative Results}

In this research, we have employed six standard evaluation metrics for categorical variables (classification) and three standard evaluation metrics for continuous variables (regression) to systematically assess the effectiveness of the proposed method. For categorical variables, we have computed precision, recall, F1-score, ROC-AUC score (Area Under the Receiver Operating Characteristic Curve), overall accuracy (micro), and mean class accuracy (macro). The confusion matrix between the predicted class and
ground truth labels was calculated first, which determined the number of four classifying conditions on pixel level: true class prediction for a given categorical variable on a positive sample (TP), false class prediction for a given categorical variable on a positive sample (FP), true class prediction for a given categorical variable on a negative sample (TN), and false class prediction for a given categorical variable on a negative sample (FN). Overall accuracy represents the percentage of correctly classified pixels (summation of TP and TN) over the total number of pixels.

\noindent \textit{Precision:} Precision is the ratio of true positive predictions to the total predicted positives. It measures the accuracy of positive predictions.
\begin{equation}
\text{Precision} = \frac{\text{True Positives}}{\text{True Positives + False Positives}}
\end{equation}

\noindent \textit{Recall:} Recall is the ratio of true positive predictions to the total actual positives. It measures the ability of the model to capture all the positive instances.
\begin{equation}
    \text{Recall} = \frac{\text{True Positives}}{\text{True Positives + False Negatives}}
\end{equation}

\noindent \textit{F1-score:} The F1-score is the harmonic mean of precision and recall. It provides a balance between precision and recall.
\begin{equation}
    \text{F1-Score} = \frac{2 \times (\text{Precision} \times \text{Recall})}{\text{Precision + Recall}}
\end{equation}

\noindent \textit{ROC-AUC Score:} ROC-AUC measures the area under the ROC curve, which is a graphical representation of the trade-off between true positive rate (sensitivity) and false positive rate (1 - specificity).

\noindent \textit{Overall Accuracy (Micro-Average)}: Micro-average accuracy considers the overall performance across all classes by aggregating individual class metrics.
\begin{dmath}
    \text{Micro Accuracy} = \frac{\text{Sum of True Positives across all classes}}{\text{Sum of (True Positives + False Positives)}}
\end{dmath}

\noindent \textit{Mean Class Accuracy (Macro-Average)}: Macro-average accuracy calculates the average accuracy for each class and then averages these values.
\begin{equation}
    \text{Macro Accuracy} = \frac{\text{Average of Class Accuracies}}{\text{Number of Classes}}
\end{equation}

Microaccuracy favors the model’s good performance for the majority class or classes, whereas macroaccuracy also gives equal weight to the minority classes. We have computed standard metrics for different categorical classes in the given Table~\ref{tab:2}. We performed the metric computation for multiple runs (at least $10$ times with different random seeds initialized) for each class to know the fluctuation of accuracy in our model. Repeatedly computing accuracy for the proposed model across multiple runs and calculating the minimum and maximum values serves several crucial purposes. Firstly, it addresses the randomness inherent in data splitting, particularly when employing techniques like random sampling or shuffling. Additionally, it accounts for the variability that certain machine-learning models may exhibit due to stochastic elements or random initialization. 

\begin{figure*}[!h]
\centering 
\subfloat[]{\includegraphics[width=0.3\textwidth]{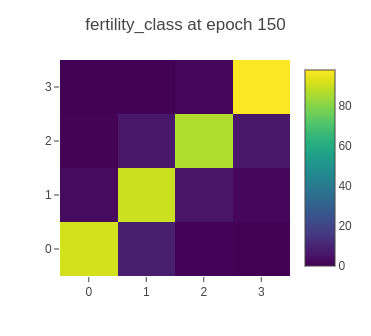}}
\subfloat[]{  \includegraphics[width=0.3\textwidth]{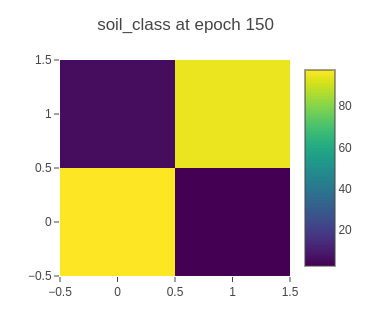}}
\subfloat[]{  \includegraphics[width=0.3\textwidth]{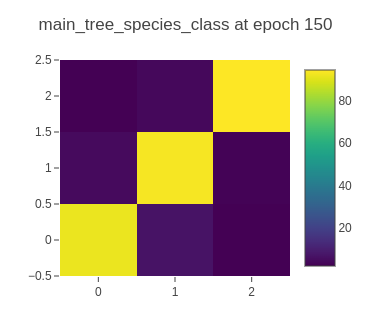}}
        
\caption{Normalized confusion matrices for the classification tasks. The horizontal axis shows the true classes, and the vertical axis shows the predicted classes}
\label{Fig9}

\end{figure*}

The practice of assessing accuracy multiple times aids in understanding the model's robustness, as extreme values may highlight sensitivity to specific conditions or data subsets. This approach also facilitates outlier detection, revealing potential issues within the dataset. When tuning hyperparameters, analyzing the range of accuracies helps gauge the model's sensitivity to different settings, informing stable configurations. Furthermore, statistical analysis, including metrics like mean class accuracy performed on multiple runs, offers insights into the central tendency and variability of the model's performance. Importantly, this method simulates real-world scenarios where the model may encounter diverse data variations, providing a more realistic assessment of its performance in different situations. From the table, it can be observed that the model's performance is consistent, which implies that the model's behavior is reproducible. It also indicates good generalization to unseen data. Furthermore, we computed normalized confusion matrices for the classification tasks at the end of the training process, as shown in figure~\ref{Fig9}. The model was evaluated on the validation set every ten epochs to monitor its learning progress. Initially, the model tried to predict the minority class due to significant repercussions from the classification loss function. However, as the training advanced and the model encountered the training samples repeatedly, its performance significantly improved by the conclusion of the training period.

\begin{table*}[]
\centering
\caption{Performance evaluation for all three different categorical variables. To ensure that the reported performance metrics are robust and representative of the model's true capability, we have conducted $10$ trials/seeds of the experiment to get min/max/mean values.}
\label{tab:2}
\renewcommand{\arraystretch}{1.2}
\resizebox{1\textwidth}{!}{%
\begin{tabular}{|l|cccccc|}
\hline
Categorical                                                  & Precision                                                               & Recall                                                                  & F1-score                                                                & \begin{tabular}[c]{@{}c@{}}ROC-AUC\\ score\end{tabular}                 & \begin{tabular}[c]{@{}c@{}}Overall Accuracy\\  (Micro)\end{tabular}     & \begin{tabular}[c]{@{}c@{}}Mean Class \\ Accuracy (Macro)\end{tabular}  \\ \hline 
Fertility Class                                              & \begin{tabular}[c]{@{}c@{}}94.66\\ Max: 94.72\\ Min: 94.50\end{tabular} & \begin{tabular}[c]{@{}c@{}}94.12\\ Max: 94.83\\ Min: 94.03\end{tabular} & \begin{tabular}[c]{@{}c@{}}94.23\\ Max: 94.79\\ Min: 94.09\end{tabular} & \begin{tabular}[c]{@{}c@{}}96.89\\ Max: 96.98\\ Min: 96.72\end{tabular} & \begin{tabular}[c]{@{}c@{}}97.03\\ Max: 97.13\\ Min: 96.96\end{tabular} & \begin{tabular}[c]{@{}c@{}}95.78\\ Max: 95.93\\ Min: 95.10\end{tabular} \\ \hline
Soil Class                                                   & \begin{tabular}[c]{@{}c@{}}98.57\\ Max: 98.63\\ Min: 98.29\end{tabular} & \begin{tabular}[c]{@{}c@{}}98.51\\ Max: 98.68\\ Min: 98.18\end{tabular} & \begin{tabular}[c]{@{}c@{}}98.53\\ Max: 98.59\\ Min: 98.21\end{tabular} & \begin{tabular}[c]{@{}c@{}}98.25\\ Max: 98.54\\ Min: 98.02\end{tabular} & \begin{tabular}[c]{@{}c@{}}98.73\\ Max: 98.49\\ Min: 98.13\end{tabular} & \begin{tabular}[c]{@{}c@{}}98.26\\ Max: 98.76\\ Min: 98.16\end{tabular} \\ \hline
\begin{tabular}[c]{@{}c@{}}Mean Tree \\ Species\end{tabular} & \begin{tabular}[c]{@{}c@{}}97.36\\ Max: 97.53\\ Min: 97.06\end{tabular} & \begin{tabular}[c]{@{}c@{}}97.29\\ Max: 97.78\\ Min: 97.08\end{tabular} & \begin{tabular}[c]{@{}c@{}}97.31\\ Max: 97.49\\ Min: 97.15\end{tabular} & \begin{tabular}[c]{@{}c@{}}98.05\\ Max: 97.96\\ Min: 97.81\end{tabular} & \begin{tabular}[c]{@{}c@{}}97.91\\ Max: 97.98\\ Min: 97.64\end{tabular} & \begin{tabular}[c]{@{}c@{}}97.55\\ Max: 97.84\\ Min: 96.98\end{tabular} \\ \hline
\end{tabular}
}
\end{table*}

For continuous class (regression method), we have considered statistical computation such as the root-mean-square error (RMSE), mean absolute error (MAE), and coefficient of determination ($R^2$) (Table~\ref{tab:3}). Among all other regression methods, RMSE is chosen for its emphasis on larger errors, providing a meaningful measure of the average magnitude of deviations in a regression model, particularly when larger errors are critical. MAE is favored for its simplicity and interpretability. It offers a straightforward average of absolute errors without emphasizing extreme values, making it suitable for scenarios where all errors are considered equally important. $R^2$ is a widely chosen metric for its comprehensive evaluation of model performance. It captures the proportion of variance explained by the model, providing a global measure of fit and predictive power. The equation for these computational methods is given as follows:

\begin{equation}
    \text{RMSE} = \sqrt{\frac{1}{n} \sum_{i=1}^{n} (y_i - \hat{y}_i)^2}
\end{equation}

\begin{equation}
    \text{MAE} = \frac{1}{n} \sum_{i=1}^{n} |y_i - \hat{y}_i|
\end{equation}

\begin{equation}
    R^2 = 1 - \frac{\sum_{i=1}^{n} (y_i - \hat{y}_i)^2}{\sum_{i=1}^{n} (y_i - \bar{y})^2}
\end{equation}

where n is the total number of data points, $y_i$ is the true value, and $\hat{y}_i$  is the predicted value for the $i^{th}$ data point. A lower RMSE indicates better precision, with values closer to $0$ suggesting a more accurate model. Similarly, smaller MAE values indicate better accuracy and reduced bias in the model. In contrast, a higher $R^2$ indicates a better fit, with $1.0$ indicating perfect prediction and $0.0$ indicating a model that does not explain any variance. In table~\ref{tab:3}, it can be observed that our proposed model demonstrates higher mean performance while maintaining lower or equivalent variability, as evidenced by reduced differences between the minimum and maximum values of RMSE, MAE, and $R^2$ when trained across $10$ different seeds. The $10$ different trials/seeds of the experiment were conducted to ensure that the reported performance metrics are robust and represent the model's true capability.

\begin{table}[h]
\centering
\caption{Performance evaluation for all ten different continuous variables.  To ensure that the reported performance metrics are robust and represent the model's true capability, we have conducted $10$ trials/seeds of the experiment to get min/max/mean values. Our proposed model demonstrates higher mean performance while maintaining lower or equivalent variability, as evidenced by reduced differences between the minimum and maximum values of RMSE, MAE, and $R^2$ when trained across 10 different seeds.}
\label{tab:3}
\resizebox{0.9\columnwidth}{!}{%
\begin{tabular}{lccc}
\toprule
Continuous                & RMSE                                                                    & MAE                                                                      & $R^2$                                                                       \\ \midrule
Basal area {[}$m^2$/ha{]}    & \begin{tabular}[c]{@{}c@{}}0.027\\Min: 0.024 \\ Max: 0.03\end{tabular}  & \begin{tabular}[c]{@{}c@{}}0.019\\Min: 0.017 \\ Max: 0.021\end{tabular}  & \begin{tabular}[c]{@{}c@{}}0.907\\ Max: 0.913\\ Min: 0.889\end{tabular} \\ \midrule
Mean dbh {[}cm{]}         & \begin{tabular}[c]{@{}c@{}}0.030\\Min: 0.027 \\ Max: 0.032\end{tabular} & \begin{tabular}[c]{@{}c@{}}0.028\\Min: 0.026 \\Max: 0.031 \end{tabular}  & \begin{tabular}[c]{@{}c@{}}0.913\\ Max: 0.927\\ Min: 0.892\end{tabular} \\ \midrule
Stem density {[}1/ha{]}   & \begin{tabular}[c]{@{}c@{}}0.021\\Min: 0.019 \\ Max: 0.025\end{tabular} & \begin{tabular}[c]{@{}c@{}}0.021\\Min: 0.018\\ Max: 0.026 \end{tabular}  & \begin{tabular}[c]{@{}c@{}}0.945\\ Max: 0.959\\ Min: 0.936\end{tabular} \\ \midrule
Mean height {[}cm{]}      & \begin{tabular}[c]{@{}c@{}}0.011\\ Min: 0.010\\ Max: 0.013\end{tabular} & \begin{tabular}[c]{@{}c@{}}0.030\\Min: 0.029 \\ Max: 0.032\end{tabular}  & \begin{tabular}[c]{@{}c@{}}0.927\\ Max: 0.938\\ Min: 0.916\end{tabular} \\ \midrule
Percentage of Pine        & \begin{tabular}[c]{@{}c@{}}0.020\\Min: 0.019 \\ Max: 0.022\end{tabular} & \begin{tabular}[c]{@{}c@{}}0.027\\Min: 0.026 \\Max: 0.028 \end{tabular}  & \begin{tabular}[c]{@{}c@{}}0.919\\ Max: 0.926\\ Min: 0.907\end{tabular} \\ \midrule
Percentage of spruce      & \begin{tabular}[c]{@{}c@{}}0.037\\Min: 0.036 \\ Max: 0.038\end{tabular} & \begin{tabular}[c]{@{}c@{}}0.036\\Min: 0.035 \\ Max: 0.038\end{tabular}  & \begin{tabular}[c]{@{}c@{}}0.926\\ Max: 0.936\\ Min: 0.918\end{tabular} \\ \midrule
Percentage of birch       & \begin{tabular}[c]{@{}c@{}}0.038\\Min: 0.037 \\ Max: 0.039\end{tabular} & \begin{tabular}[c]{@{}c@{}}0.036\\Min: 0.035 \\Max: 0. 038 \end{tabular} & \begin{tabular}[c]{@{}c@{}}0.949\\ Max: 0.958\\ Min: 0.937\end{tabular} \\ \midrule
Woody biomass             & \begin{tabular}[c]{@{}c@{}}0.026\\Min: 0.025 \\ Max: 0.027\end{tabular} & \begin{tabular}[c]{@{}c@{}}0.031\\Min: 0.030 \\ Max: 0.032\end{tabular}  & \begin{tabular}[c]{@{}c@{}}0.951\\ Max: 0.964\\ Min: 0.942\end{tabular} \\ \midrule
Leaf area index           & \begin{tabular}[c]{@{}c@{}}0.033\\Min: 0.032 \\ Max: 0.035\end{tabular} & \begin{tabular}[c]{@{}c@{}}0.028\\Min: 0.027 \\ Max: 0.029\end{tabular}  & \begin{tabular}[c]{@{}c@{}}0.916\\ Max: 0.927\\ Min: 0.908\end{tabular} \\ \midrule
Effective leaf area index & \begin{tabular}[c]{@{}c@{}}0.028\\Min: 0.027 \\ Max: 0.030\end{tabular} & \begin{tabular}[c]{@{}c@{}}0.030\\Min: 0.028\\ Max: 0.031 \end{tabular}  & \begin{tabular}[c]{@{}c@{}}0.928\\ Max: 0.931\\ Min: 0.927\end{tabular} \\ \bottomrule
\end{tabular}
}
\end{table}

\subsubsection{Qualitative Results}
\begin{figure*}[!h]
    \centering
    \includegraphics[width=\textwidth]{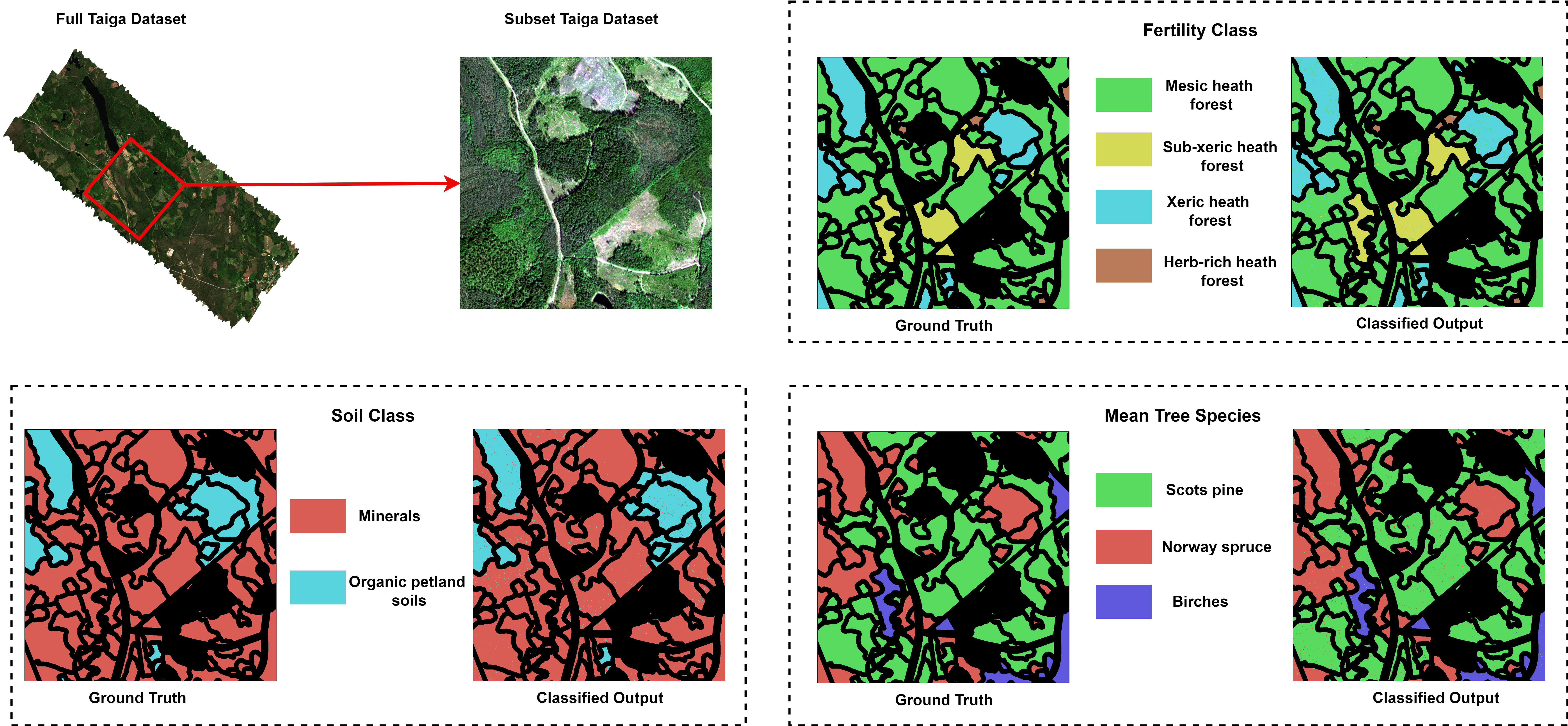}
    \caption{The pixel-wise classification map for all three categorical variables to visualize different classes of the TAIGA dataset.}
    \label{Fig10}
\end{figure*}

For the qualitative evaluation, we have a classification map for the categorical variable to visualize different classes of the TAIGA dataset, as shown in figure~\ref{Fig10}. The classification map visually represents the model's predictions, assigning specific class labels to different regions. This helps in validating the accuracy of the classification models by comparing them with ground truth data for validation and accuracy assessment. A subset image from the full TAIGA dataset has been chosen to capture all the species of different classes and ease visualization and interpretation. 

\begin{figure}[!h]
\centering 
	\subfloat[]{
        \includegraphics[width=0.45\columnwidth]{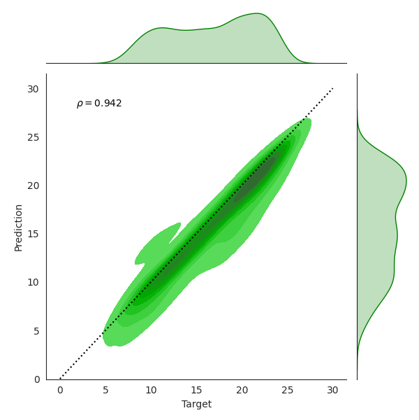}}
        \subfloat[]{  \includegraphics[width=0.45\columnwidth]{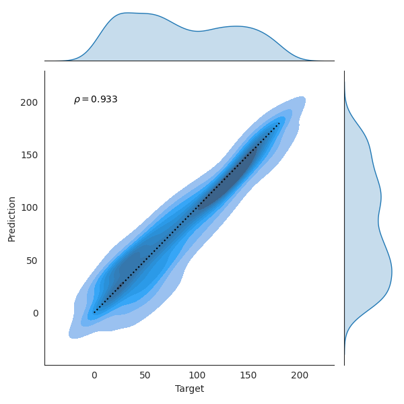}}
        
\caption{Prediction accuracy for continuous forest variable (a) Mean tree height (b) Woody biomass with Pearson correlation coefficient of $0.942$ and $0.933$ respectively.  }
\label{Fig11}

\end{figure}

\begin{figure}[!h]
\centering 
		\subfloat[]{
        \includegraphics[width=0.45\columnwidth]{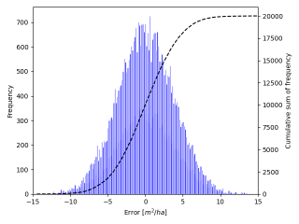}}
        \subfloat[]{  \includegraphics[width=0.45\columnwidth]{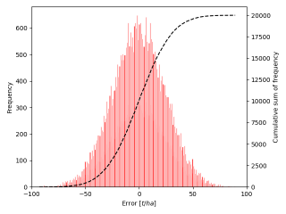}}
        
\caption{Prediction error histogram for selected continuous forest variables (a) Mean tree height (b) Woody biomass}
\label{Fig12}
\end{figure}

For the regression method, scatter plots and kernel density plots between prediction and target label have been plotted on randomly selected continuous forest variables (Figure~\ref{Fig11}). The scatter plot represents the target labels on the x-axis and the predicted labels on the y-axis. Ideally, all points should align with the line y = x or the black dotted line on the plot. The top marginal plot illustrates the true distribution, while the right marginal plot represents the predicted distribution. Additionally, kernel density estimation plots specify the density of points on the respective scatter plots, with darker regions indicating higher point concentrations. A robust model is characterized by darker regions surrounding the modes of the target distribution. The Pearson correlation coefficient is also computed to know how closely the predicted and target outputs are correlated. High Pearson correlation coefficients in the scatter plots ($\geq0.93$) prove that the prediction and target labels for all the regression tasks are strongly correlated. 
Additionally, an error histogram (Figure~\ref{Fig12}) is plotted by computing the absolute error for each test sample and plotting the histogram of the error values with 100 bins. A good model will have most of the errors falling in bins that are close to zero. The error histograms generated by our models show that the modes are in close proximity to zero. This gives an overview of how well the model performs in general. 

\subsection{Comparison Analysis}
To assess the performance of the proposed model, we have chosen four cutting-edge deep learning algorithms in the field of remote sensing, naming U-Net~\cite{ronneberger2015u}, SegNet~\cite{badrinarayanan2017segnet}, Res-U-Net~\cite{cciccek20163d}, and Deeplab V3+~\cite{chen2018encoder}, each sharing architectural similarities with our model. This ensures a fair and relevant comparison. Additionally, we've identified two methodologies that have employed identical datasets for multitask learning, aiding a precise comparison with our proposed model, naming the Mõttus model~\cite{mottus2022taiga} and the Pham model ~\cite{pham2019deep}. This allows for a more direct and accurate comparison with our proposed model. Further, we selected recent state-of-the-art networks using a transformer-based architecture for hyperspectral images, naming MSTNet~\cite{9807344} (A Multilevel Spectral–Spatial Transformer Network) and GSC-ViT~\cite{10472541} (Groupwise Separable Convolutional Vision Transformer) and a diffusion-based network naming SpectralDiff~\cite{10234379}. Below are concise summaries outlining the characteristics of each method:

\subsubsection{U-net} introduced by Ronneberger et al. ~\cite{ronneberger2015u}, is a convolutional neural network architecture notable for its U-shaped, featuring an encoder-decoder structure with skip connections to preserve spatial information. U-Net, initially designed for semantic segmentation, has been repurposed for classification and regression tasks as a base architecture for our algorithm.
\subsubsection{SegNet} is a convolutional neural network architecture specifically designed for semantic segmentation tasks, notable for its encoder-decoder structure with max-pooling indices for efficient memory usage during decoding. Introduced by Badrinarayanan et al. in 2017~\cite{badrinarayanan2017segnet}, SegNet prioritizes accurate pixel-level labeling, making it suitable for various image analysis applications, including those in the remote sensing domain.
\subsubsection{Res-U-Net} is a hybrid architecture combining the features of U-Net and residual connections from ResNet. It takes the strengths of U-Net's encoder-decoder structure for classification and feature extraction tasks while incorporating residual connections to reduce vanishing gradient issues and aid deeper network training. Introduced by Özgün Çiçek et al. in 2016 ~\cite{cciccek20163d}, Res-U-Net aims to delineate objects and features within remote sensing imagery accurately. This is particularly beneficial for tasks such as land cover classification of aerial or satellite images, where precise identification of features like vegetation and water bodies is essential. 
\subsubsection{Deeplab V3+} is an advanced image segmentation architecture notable for its encoder-decoder structure with atrous separable convolutions, aiding multi-scale feature extraction while maintaining computational efficiency. Introduced by Chen et al. in 2018 ~\cite{chen2018encoder}, it offers state-of-the-art performance in remote sensing tasks by effectively capturing spatial context and fine details in high-resolution imagery.
\subsubsection{Mõttus model} introduced the TAIGA dataset ~\cite{mottus2022taiga} where 3-D CNN pipeline was used as baseline inspired by Chen et al. ~\cite{chen2016deep}. This work used shared fully connected layers at the output to perform multitask learning tasks. We chose the same dataset for our proposed model, which has separate output layers for multitask learning that can predict discrete and continuous variables. 
\subsubsection{Pham model} uses a CNN architecture that includes seven convolution layers, one shared fully connected (FC) layer, and two FC layers for each specific task ~\cite{pham2019deep}. The depth of the network considered was seven, which is considered deep, stating the fact that most of the CNNs for HSI-related tasks are shallower. This work is considered a baseline comparison for our proposed CNN-based encoder-decoder architecture for a multitask deep learning network. 
\subsubsection{MSTNet} is a multilevel spectral-spatial transformer network for hyperspectral image classification. This network proposed by~\cite{9807344} uses an image-based classification framework to solve the inefficiency in the training and testing. They have also used a transformer encoder to extract features from the global receptive field and calculate the correlation between pixels.
\subsubsection{GSC-ViT} proposed by~\cite{10472541} is a lightweight Groupwise separable convolutional vision transformer-based deep learning network. This framework comprises a spectral calibration block, two feature extraction blocks, a global average pooling block, and a fully connected layer. Additionally, this network substitutes the multi-layer perceptron with a point-wise convolution layer to improve the residual connection structure.
\subsubsection{SpectralDiff} is a generative framework for hyperspectral image classification with diffusion models. This approach proposed by~\cite{10234379} effectively utilizes the distribution information inherent in high-dimensional and highly redundant data by iterative denoising and explicitly constructing the data generation process. This method aims to reflect the relationships between samples more accurately. The framework comprises a spectral-spatial diffusion module and an attention-based classification module.

Table~\ref{tab:4} compares the proposed model with the above-mentioned state-of-the-art methods. We evaluated classification tasks using  OA, Precision, and Recall metrics, which are suitable for assessing categorical variables. We employed RMSE and MAE as evaluation metrics for regression tasks involving continuous variables. Notably, among cutting-edge deep learning algorithms, Res-U-Net demonstrated superior performance across both classification and regression tasks, achieving an OA of $91\%$ and an RMSE of $0.038$. Compared to this, the Pham model utilizing a simple CNN architecture yields the lowest OA, followed by the Mõttus model and U-Net. It is interesting to observe that there is not much difference in OA between the proposed method and transformer and diffusion-based methods. The MSTNet and GSC-ViT have an OA of $93.83\%$ and $92.11\%$, respectively, while SpectralDiff has an OA of $92.89\%$. However, the proposed algorithm shows even higher performance levels, surpassing both classification and regression benchmarks with an OA of approximately $95\%$ and RMSE as low as $0.027$, outperforming all other evaluated models.

In the context of large hyperspectral image data, transformer networks are limited by their high computational complexity and substantial memory requirements, which are intensified by the large size of hyperspectral datasets. They also demand extensive labeled data for effective training, which is often scarce in this domain. Conversely, CNNs excel at capturing local spectral-spatial features efficiently, making them more suitable for tasks like classification and segmentation in hyperspectral imagery. Similarly, diffusion-based networks, such as diffusion probabilistic models or diffusion architectures, are primarily designed for sequential or temporal data processing and may not preserve spatial dependencies crucial in image analysis. Furthermore, diffusion models can be computationally demanding and complex to train, requiring specialized probabilistic frameworks that may not scale efficiently for large-scale hyperspectral image datasets. In contrast, CNNs are well-suited for spatial feature extraction and multitasking in image analysis tasks due to their inherent ability to capture spatial hierarchies and scalability with large image sizes. This comparison demonstrates that the proposed algorithm outperforms recent state-of-the-art methods as well as similar architecture algorithms and those utilizing the same dataset.

\begin{table}[]
\caption{Comparison with State-of-the-Art Methods: The best results (proposed model) are in bold, and the second-best are underlined.
(Higher values of OA, Precision, and Recall indicate better performance; lower values of RMSE and MAE are better). Our model shows higher average performance and lower variability across 10 trials.}
\label{tab:4}
\resizebox{1.\columnwidth}{!}{%
\begin{tabular}{lccccc}
\toprule
\multirow{2}{*}{\textbf{Method}} & \multicolumn{3}{c}{\textbf{Classification}}                                                                                                                                                                                                            & \multicolumn{2}{c}{\textbf{Regression}}                                                                                                                             \\ \cmidrule{2-6} 
                                 & OA (\%)                                                                          & Precision                                                                        & Recall                                                                           & RMSE                                                                             & MAE                                                                              \\ \midrule
U-Net                            & \begin{tabular}[c]{@{}c@{}}86.12\\ Max: 88.63\\ Min: 83.97\end{tabular}          & \begin{tabular}[c]{@{}c@{}}80.81\\ Max: 83.27\\ Min: 78.19\end{tabular}          & \begin{tabular}[c]{@{}c@{}}79.42\\ Max: 82.17\\ Min: 78.01\end{tabular}          & \begin{tabular}[c]{@{}c@{}}0.059\\ Min: 0.057\\Max: 0.061 \end{tabular}          & \begin{tabular}[c]{@{}c@{}}0.063\\Min: 0.061 \\Max: 0.065 \end{tabular}          \\ \midrule
SegNet                           & \begin{tabular}[c]{@{}c@{}}90.87\\ Max: 92.13\\ Min: 88.92\end{tabular}          & \begin{tabular}[c]{@{}c@{}}87.46\\ Max: 90.01\\ Min: 85.92\end{tabular}          & \begin{tabular}[c]{@{}c@{}}87.01\\ Max: 89.98\\ Min: 86.07\end{tabular}          & \begin{tabular}[c]{@{}c@{}}0.053\\Min: 0.052 \\Max: 0.055 \end{tabular}          & \begin{tabular}[c]{@{}c@{}}0.057\\Min: 0.056 \\ Max: 0.059\end{tabular}          \\ \midrule
Res-U-Net                        & \begin{tabular}[c]{@{}c@{}}91.02\\ Max: 92.23\\ Min: 89.78\end{tabular}          & \begin{tabular}[c]{@{}c@{}}89.17\\ Max: 90.36\\ Min: 88.25\end{tabular}          & \begin{tabular}[c]{@{}c@{}}88.79\\ Max: 89.84\\ Min: 87.07\end{tabular}          & \begin{tabular}[c]{@{}c@{}}0.038\\Min: 0.036 \\ Max: 0.039\end{tabular}          & \begin{tabular}[c]{@{}c@{}}0.034\\Min: 0.033 \\Max: 0.036 \end{tabular}          \\ \midrule
Deeplab V3+                      & \begin{tabular}[c]{@{}c@{}}89.18\\ Max: 91.01\\ Min: 88.96\end{tabular}          & \begin{tabular}[c]{@{}c@{}}84.64\\ Max: 86.78\\ Min: 83.12\end{tabular}          & \begin{tabular}[c]{@{}c@{}}85.91\\ Max: 87.21\\ Min: 83.84\end{tabular}          & \begin{tabular}[c]{@{}c@{}}0.062\\Min: 0.060 \\Max: 0.063 \end{tabular}          & \begin{tabular}[c]{@{}c@{}}0.059\\Min: 0.057 \\Max: 0.061 \end{tabular}          \\ \midrule
Mõttus Model                     & \begin{tabular}[c]{@{}c@{}}84.71\\ Max: 86.37\\ Min: 82.78\end{tabular}          & \begin{tabular}[c]{@{}c@{}}81.08\\ Max: 83.59\\ Min: 80.01\end{tabular}          & \begin{tabular}[c]{@{}c@{}}82.15\\ Max: 84.58\\ Min: 80.95\end{tabular}          & \begin{tabular}[c]{@{}c@{}}0.087\\Min: 0.085 \\Max: 0.089 \end{tabular}          & \begin{tabular}[c]{@{}c@{}}0.082\\Min: 0.081 \\Max: 0.084 \end{tabular}          \\ \midrule
Pham Model                       & \begin{tabular}[c]{@{}c@{}}82.43\\ Max: 83.96\\ Min: 81.01\end{tabular}          & \begin{tabular}[c]{@{}c@{}}79.18\\ Max: 81.48\\ Min: 78.92\end{tabular}          & \begin{tabular}[c]{@{}c@{}}80.64\\ Max: 82.37\\ Min: 79.12\end{tabular}          & \begin{tabular}[c]{@{}c@{}}0.095\\Min: 0.094 \\ Max: 0.096\end{tabular}          & \begin{tabular}[c]{@{}c@{}}0.094\\Min: 0.093 \\Max: 0.096 \end{tabular}          \\ \midrule
\underline{MSTNet}                           & \begin{tabular}[c]{@{}c@{}}\underline{93.83}\\ Max: 94.11\\ Min: 90.04\end{tabular}         & \begin{tabular}[c]{@{}c@{}}\underline{92.98}\\ Max: 93.28\\ Min: 91.26\end{tabular}          & \begin{tabular}[c]{@{}c@{}}\underline{92.14}\\ Max: 93.27\\ Min: 90.93\end{tabular}         & \begin{tabular}[c]{@{}c@{}}\underline{0.034}\\Min: 0.033 \\Max: 0.035 \end{tabular}          & \begin{tabular}[c]{@{}c@{}}\underline{0.038}\\ Min: 0.036\\ Max: 0.039\end{tabular}         \\ \midrule
GSC-ViT                          & \begin{tabular}[c]{@{}c@{}}92.11\\ Max: 93.87\\ Min: 91.08\end{tabular}          & \begin{tabular}[c]{@{}c@{}}91.27\\ Max: 93.29\\ Min: 90.24\end{tabular}          & \begin{tabular}[c]{@{}c@{}}90.79\\ Max: 91.85\\ Min: 89.27\end{tabular}          & \begin{tabular}[c]{@{}c@{}}0.046\\Min: 0.043\\  Max: 0.047\end{tabular}          & \begin{tabular}[c]{@{}c@{}}0.049\\ Min: 0.047\\ Max: 0.051\end{tabular}          \\ \midrule
SpectralDiff                     & \begin{tabular}[c]{@{}c@{}}92.89\\ Max: 93.74\\ Min: 90.69\end{tabular}          & \begin{tabular}[c]{@{}c@{}}91.73\\ Max: 93.26\\ Min: 90.86\end{tabular}          & \begin{tabular}[c]{@{}c@{}}91.48\\ Max: 93.72\\ Min: 90.21\end{tabular}          & \begin{tabular}[c]{@{}c@{}}0.039\\ Min: 0.037\\ Max: 0.041\end{tabular}          & \begin{tabular}[c]{@{}c@{}}0.040\\ Min: 0.038\\ Max: 0.042\end{tabular}          \\ \midrule
\textbf{Proposed}                & \begin{tabular}[c]{@{}c@{}}\textbf{95.92}\\ Max: 96.38\\ Min: 94.01\end{tabular} & \begin{tabular}[c]{@{}c@{}}\textbf{94.66}\\ Max: 94.72\\ Min: 94.50\end{tabular} & \begin{tabular}[c]{@{}c@{}}\textbf{94.12}\\ Max: 94.83\\ Min: 94.03\end{tabular} & \begin{tabular}[c]{@{}c@{}}\textbf{0.027}\\ Min: 0.024\\ Max: 0.003\end{tabular} & \begin{tabular}[c]{@{}c@{}}\textbf{0.019}\\Min: 0.017 \\ Max: 0.021\end{tabular} \\ \bottomrule
\end{tabular}%
}
\end{table}

\subsection{Ablation Study}

\subsubsection{Effect of different layers in the network}
Table~\ref{tab:5} illustrates the performance evaluation of the impact of processing modules of the proposed network. The bolded values denote the highest metrics achieved among the methods. With the addition of different modules to the baseline, there is an increment in OA, Precision, and Recall, accompanied by reductions in RMSE and MAE. In particular, adding a multitask loss layer yields the highest precision values and the lowest RMSE, emphasizing its significance in the architecture. Multitask loss enables simultaneous optimization of multiple objectives, thus enhancing model performance by addressing data imbalances and promoting feature learning across related tasks. The channel pooling layer also gives a significant result since the dataset is huge, and using the channel pooling layer reduces dimensionality, thus capturing important features across channels. This ablation study demonstrates the effect of different modules in a network, which helps in understanding how information is transformed throughout the network, aiding in model interpretation, optimization, and architectural improvements.

\begin{table}[]
\caption{Effect of different layers in the network. R: Residual Network; D: ATROUS CONVOLUTION; A: Attention Network; M: Multitask Loss; C: Channel Pooling}
\label{tab:5}
\resizebox{1.\columnwidth}{!}{%
\begin{tabular}{lccccc}
\toprule
\multirow{2}{*}{\textbf{Method}} & \multicolumn{3}{c}{\textbf{Classification}}                                                                                                                                                                                                            & \multicolumn{2}{c}{\textbf{Regression}}                                                                                                                             \\ \cmidrule{2-6} 
                                 & OA (\%)                                                                          & Precision                                                                        & Recall                                                                           & RMSE                                                                             & MAE                                                                              \\ \midrule
Baseline                         & \begin{tabular}[c]{@{}c@{}}79.12\\ Max: 79.87\\ Min: 78.92\end{tabular}          & \begin{tabular}[c]{@{}c@{}}82.03\\ Max: 82.78\\ Min: 81.67\end{tabular}          & \begin{tabular}[c]{@{}c@{}}80.12\\ Max: 81.02\\ Min: 79.92\end{tabular}          & \begin{tabular}[c]{@{}c@{}}0.091\\Min: 0.090 \\Max: 0.092 \end{tabular}          & \begin{tabular}[c]{@{}c@{}}0.103\\Min: 0.102 \\ Max: 0.105\end{tabular}          \\ \midrule
Baseline + R                     & \begin{tabular}[c]{@{}c@{}}84.18\\ Max: 84.76\\ Min: 83.92\end{tabular}          & \begin{tabular}[c]{@{}c@{}}84.79\\ Max: 85.28\\ Min: 84.12\end{tabular}          & \begin{tabular}[c]{@{}c@{}}85.18\\ Max: 85.93\\ Min: 84.63\end{tabular}          & \begin{tabular}[c]{@{}c@{}}0.078\\ Min: 0.077\\ Max: 0.079\end{tabular}          & \begin{tabular}[c]{@{}c@{}}0.082\\Min: 0.081 \\ Max: 0.084\end{tabular}          \\ \midrule
Baseline + RD                    & \begin{tabular}[c]{@{}c@{}}87.37\\ Max: 87.85\\ Min: 87.01\end{tabular}          & \begin{tabular}[c]{@{}c@{}}86.43\\ Max: 87.12\\ Min: 86.09\end{tabular}          & \begin{tabular}[c]{@{}c@{}}86.01\\ Max: 86.92\\ Min: 85.38\end{tabular}          & \begin{tabular}[c]{@{}c@{}}0.057\\Min: 0.057 \\Max: 0.058 \end{tabular}          & \begin{tabular}[c]{@{}c@{}}0.061\\Min:0.060 \\ Max: 0.062\end{tabular}           \\ \midrule
Baseline + RDA                   & \begin{tabular}[c]{@{}c@{}}90.49\\ Max: 91.13\\ Min: 89.93\end{tabular}          & \begin{tabular}[c]{@{}c@{}}89.78\\ Max: 90.34\\ Min: 89.21\end{tabular}          & \begin{tabular}[c]{@{}c@{}}89.37\\ Max: 90.02\\ Min: 89.11\end{tabular}          & \begin{tabular}[c]{@{}c@{}}0.042\\ Min: 0.041 \\Max: 0.043\end{tabular}          & \begin{tabular}[c]{@{}c@{}}0.047\\Min: 0.045 \\ Max: 0.048\end{tabular}          \\ \midrule
Baseline + RDAM                  & \begin{tabular}[c]{@{}c@{}}91.87\\ Max: 92.34\\ Min: 91.27\end{tabular}          & \textbf{\begin{tabular}[c]{@{}c@{}}95.02\\ Max: 95.78\\ Min: 94.82\end{tabular}} & \begin{tabular}[c]{@{}c@{}}90.64\\ Max: 91.27\\ Min: 90.92\end{tabular}          & \textbf{\begin{tabular}[c]{@{}c@{}}0.027\\Min: 0.026 \\ Max: 0.028\end{tabular}} & \begin{tabular}[c]{@{}c@{}}0.032\\Min: 0.031 \\ Max: 0.033\end{tabular}          \\ \midrule
Baseline + DAC                   & \begin{tabular}[c]{@{}c@{}}91.13\\ Max: 91.89\\ Min: 90.94\end{tabular}          & \begin{tabular}[c]{@{}c@{}}91.34\\ Max: 91.93\\ Min: 90.72\end{tabular}          & \begin{tabular}[c]{@{}c@{}}90.98\\ Max: 91.34\\ Min: 90.47\end{tabular}          & \begin{tabular}[c]{@{}c@{}}0.033\\Min: 0.032 \\ Max: 0.034\end{tabular}          & \begin{tabular}[c]{@{}c@{}}0.030\\ Min: 0.029\\Max: 0.031 \end{tabular}          \\ \midrule
Proposed                         & \textbf{\begin{tabular}[c]{@{}c@{}}95.92\\ Max: 96.38\\ Min: 94.01\end{tabular}} & \begin{tabular}[c]{@{}c@{}}94.66\\ Max: 94.72\\ Min: 94.50\end{tabular}          & \textbf{\begin{tabular}[c]{@{}c@{}}94.12\\ Max: 94.83\\ Min: 94.03\end{tabular}} & \begin{tabular}[c]{@{}c@{}}0.027\\Min: 0.024 \\Max: 0.003 \end{tabular}          & \textbf{\begin{tabular}[c]{@{}c@{}}0.019\\Min: 0.017 \\Max: 0.021 \end{tabular}} \\ \bottomrule
\end{tabular}%
}
\end{table}

\subsubsection{Effect of the skip, batch normalization (BN), and global average pooling (GAP) layer}

\begin{figure}
    \centering
    \includegraphics[width=1\linewidth]{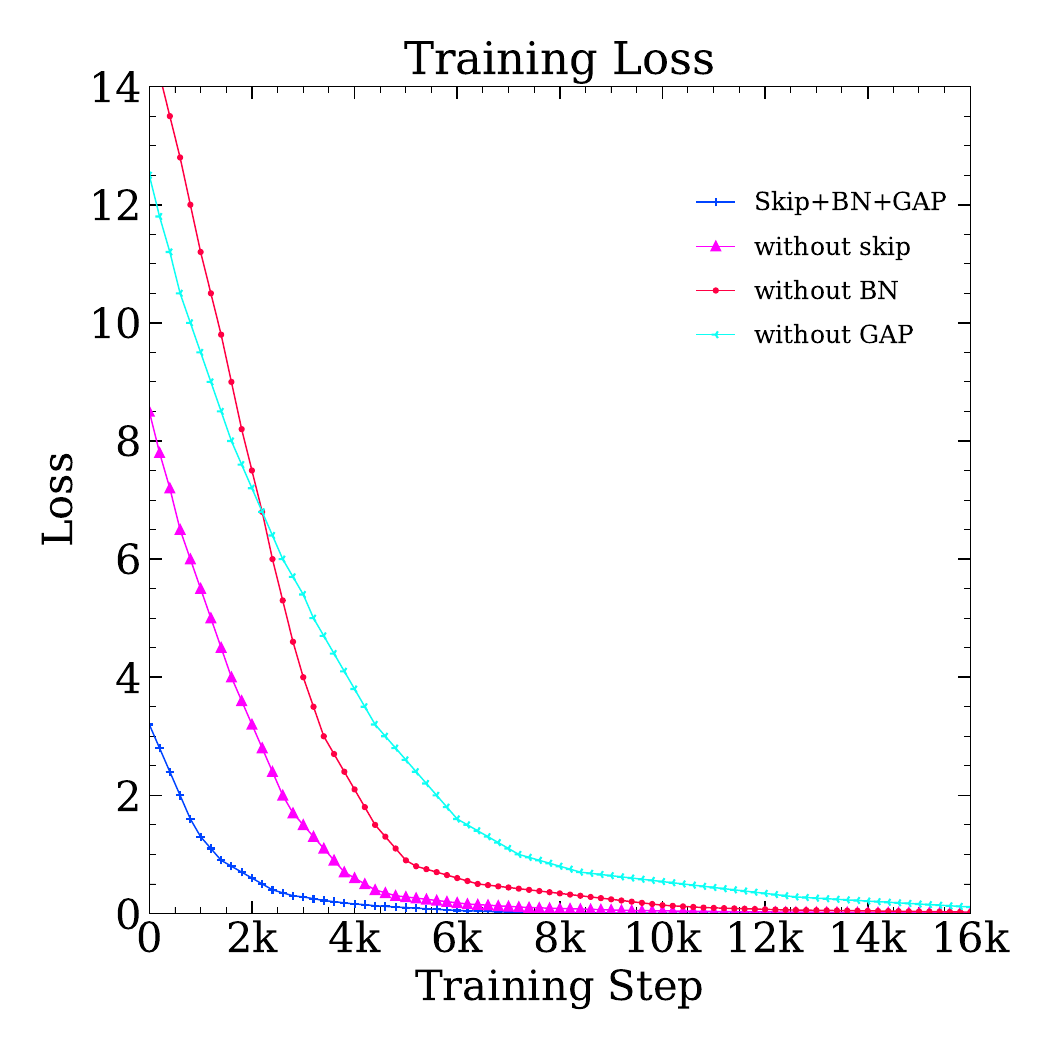}
    \caption{Effect of skip connections, batch normalization (BN), and global average pooling (GAP) layer on the proposed method}
    \label{Fig13}
\end{figure}

The impact of skip, BN, and GAP layers is studied through training loss, assessing how the proposed model converges with and without these layers. The analysis, as shown in figure~\ref{Fig13}, reveals that the combined utilization of all three options notably enhances the convergence of the training loss curve compared to their individual usage. These layers are recognized as effective regularizers for neural networks and are widely embraced by the deep learning community for enhancing model performance and generalization. Skip layers aid in the training of deep networks and reducing vanishing gradient problems, while batch normalization ensures stable and accelerated training by normalizing activations within each mini-batch, improving convergence and regularization. The global average pooling layer condenses feature maps into a single vector, reducing model complexity and overfitting while preserving spatial information. The skip layer shows its dominance over BN and GAP by showing a significant contribution to the error reduction for this dataset. Thus, including all three layers is essential for accelerating the learning process by effectively reducing the training error.

\subsubsection{Effect of cost-sensitive and focal loss}
In our proposed approach, we have incorporated cost-sensitive learning and focal loss to tackle class imbalance issues effectively. For cost-sensitive learning, the inverse median frequency method was applied to calculate class weights, which applies strict consequences on errors from minority class samples compared to majority class samples. This weighting scheme was also utilized for the cross-entropy term in focal loss. It can be observed from table~\ref{tab:6} that focal loss without the class weights for the cross-entropy loss performs poorly compared to the cost-sensitive learning approach. Therefore, we have incorporated class weights into the focal loss to address data imbalance in our study, ensuring the model is not biased towards the majority classes.

\begin{table}[]
\centering
\caption{Effect of cost-sensitive and focal loss with and without class weights on classification and regression evaluation metrics.}
\label{tab:6}
\resizebox{1.\columnwidth}{!}{%
\begin{tabular}{lccccc}
\toprule
\multirow{2}{*}{\textbf{Method}} & \multicolumn{3}{c}{\textbf{Classification}}      & \multicolumn{2}{c}{\textbf{Regression}} \\ \cmidrule{2-6} 
                                 & OA             & Precision      & Recall         & RMSE               & MAE                \\ \midrule
Cost-Sensitive                   & 87.23          & 85.26          & 86.81          & 0.073              & 0.078              \\
Focal Loss without class weights & 80.72          & 79.10          & 80.23          & 0.093              & 0.098              \\
Focal Loss with class weights    & \textbf{90.09} & \textbf{89.63} & \textbf{89.12} & \textbf{0.051}     & \textbf{0.057}     \\ \bottomrule
\end{tabular}%
}
\end{table}

\subsubsection{Evaluation of Complexity in the Proposed Multitask Model}
We conducted a comparative analysis with state-of-the-art deep learning methods on our dataset to evaluate the trade-off between performance and complexity in our proposed multitask model. Table~\ref{tab:7} summarizes key metrics, including model parameters, floating point operations (FLOPs), training epoch duration, and overall accuracy. Among the models evaluated, SpectralDiff shows the highest complexity, characterized by a large parameter count and extensive training time required to achieve satisfactory results. In contrast, the U-Net model, despite its simpler architecture with fewer parameters and shorter training time, demonstrated average performance accuracy. Furthermore, while the MSTNet showcased the second-highest accuracy (underlined in table~\ref{tab:4}), its substantial parameter size and increased FLOPs resulted in prolonged training times, making it less practical for large-scale hyperspectral datasets.
On the other hand, our proposed multitask model has the lowest parameter count as it utilizes shared encoder layers and an ASPP module with varied dilation rates. This significantly reduces the parameter count compared to its single-task counterpart and other state-of-the-art methods. This architectural choice not only minimized training time but also enhanced efficiency in capturing multi-scale contextual information without the need for additional parameters. To provide a comprehensive evaluation, we report the total computing time spent on evaluating each method, considering the initial conditions and variations in computational resources. Our method's total evaluation time was 42.8 hours (training time per epoch $\times$ number of epoch), while the second-best OA performance model, MSTNet's total evaluation time, was 88.95 hours. This indicates that the proposed multitask model emerged as a lightweight yet effective solution, striking a balance between model complexity, training efficiency, and performance accuracy suitable for large-scale hyperspectral data analysis.


\begin{table}[]
\centering
\caption{Evaluation of complexity in the proposed multitask model compared to State-of-the-Art methods.}
\label{tab:7}
\resizebox{1.\columnwidth}{!}{%
\begin{tabular}{lcccc}
\toprule
Method                                                                    & \begin{tabular}[c]{@{}c@{}}Number of\\ Parameters\\ (M)\end{tabular} & \begin{tabular}[c]{@{}c@{}}FLOPs \\ (G)\end{tabular} & \begin{tabular}[c]{@{}c@{}}Training Time\\ per epoch\\ (sec)\end{tabular} & \begin{tabular}[c]{@{}c@{}}OA\\ (\%)\end{tabular} \\ \midrule
U-Net                                                                     & 28.93                                                                & 112.67                                               & \textbf{112.83}                                                           & 86.12                                             \\ 
SegNet                                                                    & 51.44                                                                & 160.41                                               & 1219.48                                                                   & 90.87                                             \\ 
Res-U-Net                                                                 & 44.9                                                                 & 71.68                                                & 474.82                                                                    & 91.02                                             \\ 
Deeplab V3+                                                               & 53.6                                                                 & 82.7                                                 & 305.98                                                                    & 89.18                                             \\ 
Mõttus Model                                                              & 37.82                                                                & 61.28                                                & 183.42                                                                    & 84.71                                             \\ 
Pham Model                                                                & 41.37                                                                & \textbf{32.48}                                       & 247.39                                                                    & 82.43                                             \\ 
MSTNet                                                                    & 54.28                                                                & 102.46                                               & 2134.92                                                                   & 93.83                                             \\ 
GSC-ViT                                                                   & 39.19                                                                & 63.87                                                & 1948.67                                                                   & 92.11                                             \\ 
SpectralDiff                                                              & 61.94                                                                & 121.47                                               & 2897.67                                                                   & 92.89                                             \\ 
\begin{tabular}[c]{@{}c@{}}Proposed as Single-task\end{tabular}        & 37.32                                                                & 78.14                                                & 4593.67                                                                   & 91.79                                             \\ 
\textbf{\begin{tabular}[c]{@{}c@{}}Proposed as Multitask\end{tabular}} & \textbf{19.68}                                                       & 41.79                                                & 1028.76                                                                   & \textbf{95.92}                                    \\ \bottomrule
\end{tabular}%
}
\end{table}

\section{Conclusion}
In this article, we proposed a novel multitask deep-learning model for concurrent classification and regression of hyperspectral images. We have considered the TAIGA dataset, a larger hyperspectral dataset that consists of 13 forest variables, among which are three categorical (with different sub-classes) and ten continuous variables (biophysical parameters). We considered sharing encoder and task-specific decoder architecture in the proposed framework. Additionally, a dilated convolutional block called dense ASPP was considered to capture multi-scale contextual information effectively. Further, considering the importance of both spectral and spatial information, we implemented a spectra-spatial attention block to enable selective information processing where task-specific features were prioritized, improving generalization. We also integrated cost-sensitive learning and focal loss techniques to address the class imbalance issue and achieve task balancing in the proposed multitask framework. We aimed to improve model performance and efficiency across various tasks by computing multitask loss and optimizing the model parameters within this framework. Extensive experiments conducted on a large hyperspectral dataset showcase the superiority of the proposed method compared to both classical and state-of-the-art models. Ablation studies further validate the effectiveness of different parameters used for this work. Further work is being taken up where the model can be tuned to learn more tasks on different datasets for HSI classification and reduce the computation time by an iterative process, thus increasing algorithm efficiency.

\bibliographystyle{IEEEtran}
\bibliography{ref}
\end{document}